\documentclass[journal]{IEEEtran}

\usepackage{times}

\usepackage[numbers]{natbib}
\usepackage{multicol}
\usepackage[bookmarks=true]{hyperref}

\usepackage[table,xcdraw]{xcolor}
\usepackage{graphicx} 
\usepackage{amsmath,amssymb}
\usepackage{nth}
\usepackage{color} 
\definecolor{mygreen}{RGB}{28,172,0} 
\definecolor{mylilas}{RGB}{170,55,241}
\usepackage{makecell}
\usepackage[utf8]{inputenc}
\usepackage{glossaries}
\usepackage{subcaption}
\usepackage[export]{adjustbox}
\usepackage{multirow}
\usepackage{svg}
\usepackage{wrapfig}
\usepackage{mathtools}

\usepackage[linesnumbered,ruled,vlined]{algorithm2e}

\SetCommentSty{mycommfont}

\def\HiLi{\leavevmode\rlap{\hbox to \hsize{\color{yellow!50}\leaders\hrule height .8\baselineskip depth .5ex\hfill}}}

\DeclareMathOperator*{\argmax}{arg\,max}
\DeclareMathOperator*{\argmin}{arg\,min}

\newcommand{\rev}[1]{#1}

\usepackage{lipsum}
\usepackage{float}
\makeatletter
\def\BState{\State\hskip-\ALG@thistlm}
\makeatother
\newcommand{\norm}[1] {||#1||}
\usepackage{tabularx,ragged2e,booktabs,caption}

\newcolumntype{C}[1]{>{\Centering}m{#1}}

\newcommand{\normal}{\mathcal{N}}

\usepackage{flexisym}
\newcommand{\BEAS}{\begin{eqnarray*}}
\newcommand{\EEAS}{\end{eqnarray*}}
\newcommand{\BEA}{\begin{eqnarray}}
\newcommand{\EEA}{\end{eqnarray}}
\newcommand{\BEQ}{\begin{equation}}
\newcommand{\EEQ}{\end{equation}}
\newcommand{\BIT}{\begin{itemize}}
\newcommand{\EIT}{\end{itemize}}

\newcommand{\st}{\text{s.t.}}

\newcommand{\reals}{\mathbb{R}}

\newcommand{\diag}{\mathop{\textbf{diag}}}

\newcommand{\Expect}{\mathop{\mathbb{E}}}
\newcommand{\var}{\mathop{\textbf{var}}} 

\newcommand{\sign}{\mathop{\textbf{sign}}}

   {\begin{list}{}{%
    \setlength{\rightmargin}{0\linewidth}%
    \setlength{\leftmargin}{.05\linewidth}}%
    \sffamily\small
    \item[]{\setlength{\parskip}{0ex}\hrulefill\par%
    \nopagebreak{}}}%
   {{\setlength{\parskip}{-1ex}\nopagebreak\par\hrulefill} \end{list}}

\graphicspath{{media/}}
\usepackage{mydefs}

\begin{document}

\title{\rumiabv{}: Rummaging Using Mutual Information}


\author{Sheng Zhong$^{1}$, Nima Fazeli$^{1}$, and Dmitry Berenson$^{1}$%
\thanks{$^{1}$Department of Robotics, University of Michigan, MI 48109, USA
        {\tt\footnotesize \{zhsh, nfz, dmitryb\}@umich.edu}}%
}



%

\maketitle

\begin{abstract}

This paper presents \ruminame{} (\textbf{\rumiabv{}}), a method for online generation of robot action sequences to gather information about the pose of a known movable object in visually-occluded environments. Focusing on contact-rich rummaging, our approach leverages mutual information between the object pose distribution and robot trajectory for action planning. From an observed partial point cloud, \rumiabv{} deduces the compatible object pose distribution and approximates the mutual information of it with workspace occupancy in real time. Based on this, we develop an information gain cost function and a reachability cost function to keep the object within the robot's reach. These are integrated into a model predictive control (MPC) framework with a stochastic dynamics model, updating the pose distribution in a closed loop. Key contributions include a new belief framework for object pose estimation, an efficient information gain computation strategy, and a robust MPC-based control scheme. RUMI demonstrates superior performance in both simulated and real tasks compared to baseline methods.

\end{abstract}

\IEEEpeerreviewmaketitle

\section{Introduction}

Active exploration, the process of autonomously planning actions to gather more information about a target quantity, is a core problem in robotics, particularly when dealing with unknown environments~\cite{bajcsy2018revisiting}. This problem encompasses a range of scenarios, differentiated by the type of robot (e.g., mobile vs. stationary), the primary sensor modality (often vision), and the specific quantity to be estimated.

As robotics applications have transitioned from known, structured environments like factories to the unknown, dynamic environments of homes, new challenges have emerged. One critical application area is object manipulation, where visual perception is often hindered by occlusions caused by both the environment and the objects themselves~\cite{zhong2023chsel}. To address these challenges, we focus on actively exploring to estimate the pose of a movable object with a known shape through contact-rich interactions, commonly referred to as rummaging.

\begin{figure}[ht!]
    \centering
    \includegraphics[width=\linewidth]{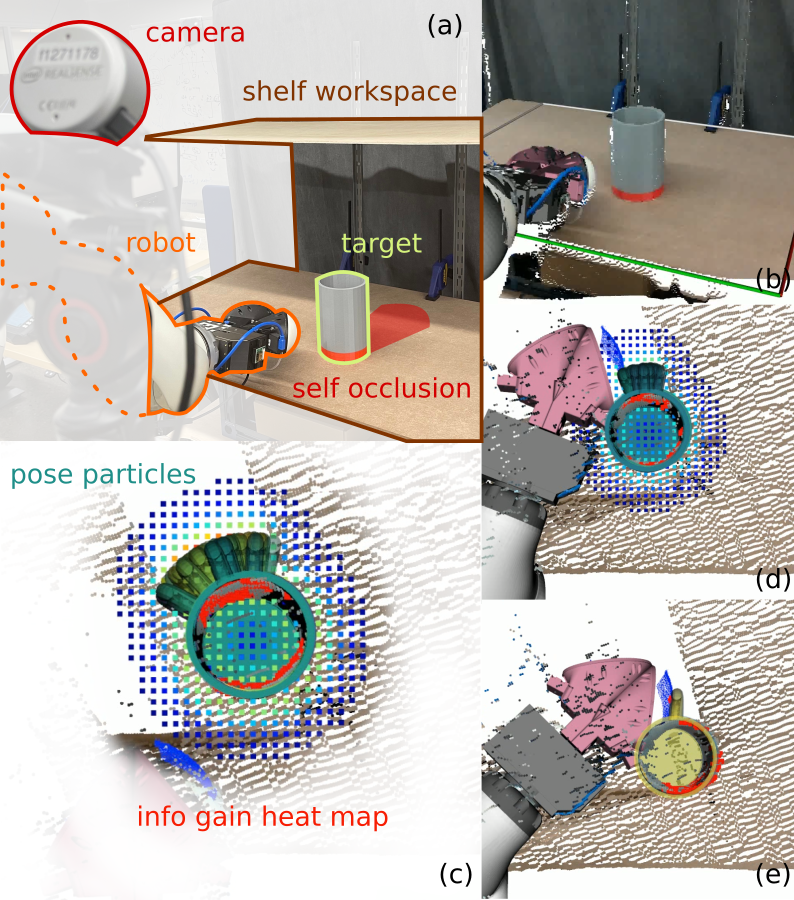}
    \caption{(a) A real-world active exploration experiment where the goal is to estimate the pose of a movable mug. The mug pose is ambiguous due to self occlusion. (b) The initial point cloud view of the scene from the camera perspective. (c) \rev{Top-down view:} \rumiabv{} maintains a belief over the mug's pose using a particle filter, where the pose particles are shown as overlaid objects with their relative likelihood indicated by color ranging from blue to yellow.
    Observed surface points are in red. From the pose particles and observations, \rumiabv{} generates an \infofieldname{} to plan over, shown as a heat map. \rev{Note the ambiguity and resulting information gain is concentrated on where the handle could be.} (d) Even without making contact with the handle, the robot sweeps out free space \rev{eliminating pose particles with handles on the left side and constraining the pose.} (e) Finally, making contact produces an accurate pose estimate.}
    \label{RUMI:fig:teaser}
\end{figure}

Occlusions of the target object, both from itself and from other objects, motivate the need to use contact to determine the object's pose. Our prior work has investigated how to track the position of contact points during rummaging with an unknown number of objects~\cite{zhong2022soft}, and how to estimate the plausible set of object poses given observed contact and free space points~\cite{zhong2023chsel}. However, the problem of how to plan information-gathering trajectories to estimate a movable object's pose is still under-explored. 
A primary challenge is the object's mobility, coupled with the requirement for contact-based information collection. Without careful planning, making contact can inadvertently push the object out of the robot's workspace, as evidenced in our experiments.

Active exploration is often framed from an information-theoretic perspective, where the quantity to be estimated is treated as a random variable, and actions are selected to minimize its uncertainty. This approach can be computationally expensive, necessitating a trade-off between accuracy and speed or limiting the exploration to a single next best action. Additionally, some methods restrict the action space to movements along the object’s surface~\cite{suresh2022midastouch},~\cite{driess2017active}. While this restriction simplifies the problem, it also limits the robot’s capabilities. Instead, we aim to enable robots to make and break contact dynamically throughout the rummaging process, enhancing their exploratory capabilities.


To address the above challenges, we present \ruminame{} (\textbf{\rumiabv{}}), an active exploration method. Specifically, our contributions include:
\begin{enumerate}
    \item a framework for creating and updating a belief over poses given observed point clouds, augmented with volumetric semantics such as whether each point is in free space or on the surface of the object, based on the discrepancy formulated in CHSEL~\cite{zhong2023chsel}
    \item a measure of information gain based on the mutual information between the object pose and volumetric semantics at the positions that the robot trajectory will cover, and show that it can be efficiently computed in parallel for dense workspace points in real time
    \item a closed loop MPC planning framework using cost functions based on the information gain and maintaining object reachability, and a stochastic object dynamics model
\end{enumerate}
In our experiments, we show that \rumiabv{} is the only method to achieve consistent success in simulated and real robot rummaging tasks across various objects.


\section{Related Work}
In a broad sense, we focus on the problem of actively exploring an unknown environment to reduce the uncertainty of some quantity. 
\rev{
We specifically consider the distinction of active perception, or sensor path planning~\cite{ryan2010particle},~\cite{cai2009information},~\cite{bajcsy2018revisiting}, in which a robot explores an environment without interacting and changing it, with interactive perception~\cite{bohg2017interactive}, in which the robot must change the environment to explore it.
}
This paper focuses on \rev{the interactive perception problem of} estimating the pose of a movable rigid object with a known shape \rev{using a robot arm}.
\rev{
The environments we are interested in include cabinets and other tight spaces which limit the ability of the robot arm to grasp top-down. Purely vision-based approaches in these environments suffer from occlusion, a restricted field of view, and the close range of the robot wrist to objects during the course of rummaging (making it difficult to use a wrist-camera). Additionally, we consider tasks that require physically moving other objects in order to access the target object in Section~\ref{RUMI:sec:clutter}.
}

In general, active exploration is the iterative process of:
\begin{enumerate}
    \item forming a belief over state given observations
    \item computing expected information gain over a workspace
    \item planning an action sequence
    \item executing some of the action sequence and collecting observations
\end{enumerate}

\subsection{Representing Belief}
Representations suitable for active exploration have been studied extensively. In many cases, parametric filters like the extended Kalman Filter (EKF)~\cite{sim2005global},~\cite{leung2006planning} may be used when the posterior of the quantity of measure should be approximately Gaussian. Otherwise, non-parametric methods like particle filters~\cite{deng2021poserbpf},~\cite{koval2015pose} are often used. Occupancy grids have also been popular, e.g. used in the simultaneous localization and mapping (SLAM) variant of active exploration~\cite{meyer2012occupancy},~\cite{vespa2018efficient},~\cite{carrillo2015autonomous}. In particular, when assuming each grid cell is independent, information gain based on the entropy of all the cells may be efficiently computed on an occupancy grid. We make a similar assumption that enables efficient computation of our information gain. 

Recently, Gaussian processes (GPs)~\cite{jadidi2015mutual}  have also been used for estimating object shape. GP implicit surfaces (GPIS) have shown strong representation power~\cite{dragiev2011gaussian},~\cite{driess2017active}. GPIS uses a GP to output a field in which the 0-level set represents the surface of the object. In our method, we do not need the full representation power of a GP since we have a known object shape. Instead, we use a particle filter to represent the pose distribution, and present a novel way to evaluate the particle probabilities given an observed point cloud. 

\subsection{Information Gain}
The information gain can be formulated in many ways, often depending on the belief representation. For GPIS the variance of the GP~\cite{driess2017active}, or the differential entropy of the GP for adding a new data point~\cite{driess2019active} can be evaluated directly and used. However, despite work on geometric shape priors for GPIS~\cite{martens2016geometric}, there remains no satisfactory way to condition a GP on a known shape with unknown pose. We implement a GPIS baseline and condition it on the shape by augmenting the input data. Mutual information between observations and the estimated quantity is also common~\cite{jadidi2015mutual},~\cite{macdonald2019active}, which measures the reduction in uncertainty of the estimated quantity given the observations. Thus, we formulate our information gain function based on the mutual information between the object pose and the occupancy at points a robot trajectory would sweep out.

\subsection{Planning}
Searching for an optimally-informative trajectory is usually computationally intensive. GP-based methods in particular are limited by inference times that grow rapidly with increasing number of data points, often addressed by using sparse GPs or downsampling to trade off accuracy~\cite{snelson2005sparse}. Some methods greedily selects the optimal next configuration, and additionally constrain the action space to slide along the surface of the object~\cite{suresh2022midastouch}, ~\cite{driess2017active}. 
Our formulation of the information gain allows us to efficiently evaluate it for many query trajectories in parallel, enabling us to use longer-horizon planning methods such as sampling-based model predictive control in a closed loop. We consider difficult tasks which necessitate long horizon planning.

Active exploration problems also differs by sensing modality. In the context of object shape and pose estimation, the most common modality is visual perception, with the common framing of the problem as finding the next best view~\cite{krainin2011autonomous}. Tactile approaches have also demonstrated success~\cite{yi2016active},~\cite{driess2017active}, as well as hybrid approaches~\cite{rustler2022active},~\cite{smith2021active}. Tightly coupled with sensing modality is the distinction of whether the robot is passively observing the environment or actively interacting with and changing the environment as in the interactive perception problem~\cite{bohg2017interactive}. \rumiabv{} is a hybrid approach for interactive perception, primarily relying on contact-rich interactions using tactile sensors, but also leveraging visual perception to initialize pose estimates. 
Unlike most other methods for object pose or shape estimation, we do not assume the object is stationary, which accounts for a large part of the difficulty. 
\rev{
We additionally consider an unknown number of unknown objects cluttered around the target object in Section~\ref{RUMI:sec:clutter}.
}
The closest method to ours is Act-VH~\cite{rustler2022active}, which trains an implicit surface neural network to output hypothesis voxel grids of seen objects given a partially observed point cloud and selects the best point to probe next. One major weakness of this method is the need to either retrain their network on all candidate objects whenever there is a new target object, or to train a network per object and assume object identity is known. Our method can be applied to new known objects without any training. Additionally, their object is in between the robot and the camera, meaning that the visually-occluded region is highly reachable, bypassing a major challenge that we address. Lastly, we consider the information gain from full robot trajectories rather than a single next point to probe.

\section{Problem Statement}

Let $\config \in \reals^\nconfig$ denote the robot configuration, and $\us \in \reals^\nus$ denote control. 
We study a single robot exploring an unmodeled environment, using limited visual perception and contact-heavy rummaging to estimate the pose of a single movable rigid target object 
of known shape.
A rigid object's configuration is defined by its pose, a transform $\T \in \poseset$. Every $\T$ can be identified with a $\reals^{4\times4}$ homogeneous transformation matrix, and for convenience, we use $\xtrans = \T\x$ to denote the homogeneous transform of point $\x \in \reals^3$ from world frame coordinates to the object frame of $\T$ (homogeneous coordinates have 1 appended).
There is an underlying dynamics function $\configdyn: \reals^\nconfig \times \reals^\nus \rightarrow \reals^\nconfig$ that we do not know, but are given the free space dynamics function $\configdyngiven: \reals^\nconfig \times \reals^\nus \rightarrow \reals^\nconfig$. The difference in dynamics is primarily due to contact between the robot and the target object. We are interested in generating a fixed length trajectory of $\nsteps$ actions, $\us_1,...,\us_{\nsteps}$ to actively explore and estimate the target object's pose.

Specifically, we have the target object's precomputed object frame signed distance function (SDF) derived from its 3D model, $\sdf: \reals^3 \rightarrow \reals$. 
After each action, sensors observe a set of points at time $t:$ $\xthis_t = \{(\x_1,\s_1),...,(\x_\nobs, \s_\nobs)\}_t$ with observed world positions $\x_\n \in \reals^3$ and semantics $\s_\n$ (described below). For convenience, we refer to a pair of position and semantics as a \textit{\descriptor{}}. Let $\xall_t$ denote the accumulated set of \descriptor{}s up to and including time $t$. Sensors may include but are not limited to robot proprioception, end-effector mounted tactile sensors, and external cameras, 
\rev{
but all must produce outputs convertible to \descriptor{}s.
}

 We treat the pose of the target object as a random variable and define $p(\T | \xall_t)$ as the posterior probability distribution over poses given $\xall_t$. Observation noise, object symmetry, and the partial nature of $\xall_t$ results in pose uncertainty.

\begin{figure}
    \centering
    \includegraphics[width=\linewidth]{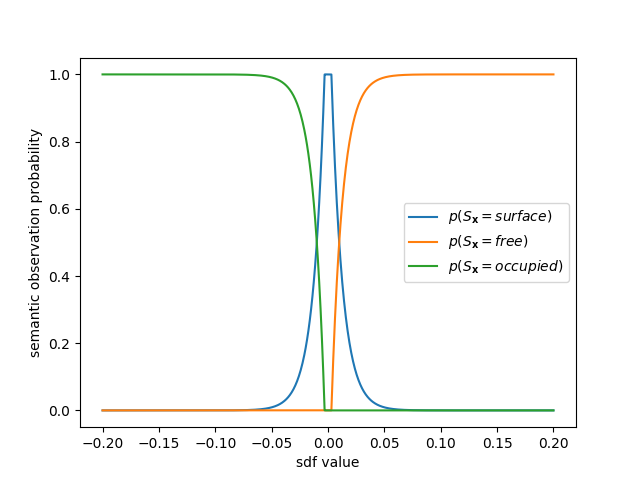}
    \caption{Example sensor model that gives a probability of observing each semantics class given a SDF value.}
    \label{RUMI:fig:sensor}
\end{figure}

Let $\T^*$ be the true object transform, then the observed semantics are
\begin{align*}
\s_\n = \begin{cases} \free & \text{implies $\sdf(\T^*\x_\n) > 0$} \\
\occupied & \text{implies $\sdf(\T^*\x_\n) < 0$} \\
\surface & \text{implies $\sdf(\T^*\x_\n) = 0$} \end{cases}
\end{align*}
For a workspace point $\x$ that we have not observed, its semantics is a discrete random variable $\Sx$ with the shorthand $p(\Sx)=p(\s|\x)$. 
We are given a sensor model $p(\Sx | \T) = p(\Sx | \sdf(\T\x))$ such as in Fig.~\ref{RUMI:fig:sensor} that gives the probability of observing each $\s$ value given a SDF value. 
The sensor model does not consider uncertainty over the position, and we assume we are given exact positions with only uncertainty over semantics $\Sx$.

Given a prior $p(\T)$, and starting at $\config_1$, our goal is to estimate the pose of the object by maximizing the expected information gain after $\nsteps$ actions:

\begin{equation}\label{RUMI:eq:mi objective}
    \begin{array}{cl}
    \argmax\limits_{\us_1,...,\us_{\nsteps}}&\Expect_{\xall_\nsteps}[D_{KL}( p(\solution | \xall_\nsteps ) || p(\solution) )]\\ 
    \st&\config_{t+1} = \configdyn(\config_t, \us_t), \ 
    t=1,...,\nsteps\\
    \end{array}
\end{equation}

The expectation is over the semantics of each position in $\xall_\nsteps$. Note that this is equivalent to the mutual information between $\T$ and $\xall_\nsteps$, $I(\T ; \xall_\nsteps)$~\cite{murphy2012machine}. 

The challenge of this problem comes from the need for contact-based perception due to limited sensing capabilities, coupled with the fact that the target object is movable. Moreover, an ineffective action sequence can result in undesirable contacts, potentially pushing the object out of the robot's reachable workspace.

We evaluate the quality of the estimated pose distribution by evaluating the likelihood of the ground truth pose $\likelihood(\Top|\xallt)$, or equivalently its 
\rev{
\textit{negative log likelihood} (NLL). Low NLL indicates both certainty and correctness of the pose distribution. 
}
We do so by sampling a set of surface points in the object frame and transforming them by $\Top$ to produce world positions $\X$.  We then evaluate the NLL of all of the points having $\surface$ semantics:
\begin{align}\label{RUMI:eq:nll}
    \nll(\xall) &= - \log p(\bigcap_{\x \in \X} \Sx = \surface | \xall) 
\end{align}

We use this metric as well as computational efficiency to evaluate our method against baselines and ablations.
\section{Method}
\begin{figure*}[h!]
    \centering
    \includegraphics[width=\linewidth]{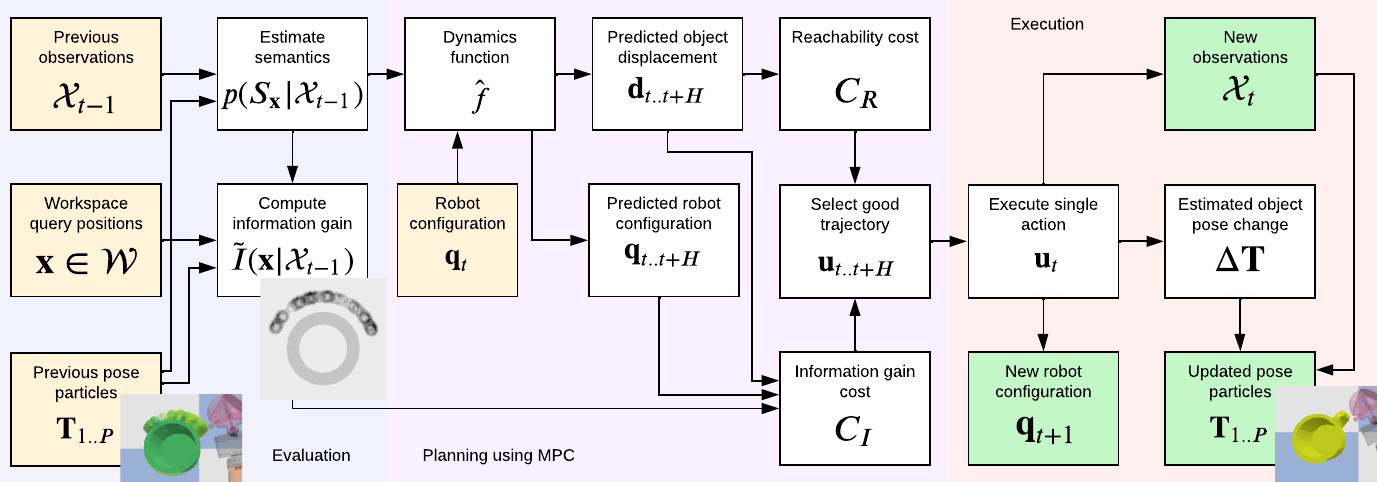}
    \caption{Flow chart showing one time step of \rumiabv{}'s approach for solving Eq.~\ref{RUMI:eq:mi objective}. Beige blocks are inputs to this time step while green ones are outputs of this time step. The process is also into evaluating current information gain, planning into the future using the information gain, and executing one step of the plan and updating observations.}
    \label{RUMI:fig:highlevel}
\end{figure*}
Our high level approach to addressing the problem in Eq.~\ref{RUMI:eq:mi objective} is depicted in Fig.~\ref{RUMI:fig:highlevel}. We represent the pose posterior $p(\T | \xall)$ with a particle filter and describe how to evaluate $p(\T | \xall)$. Next, we present a tractable surrogate for information gain that we develop into a cost function for model predictive control (MPC). To discourage trajectories that move the target object out of the robot's reachable area, we develop an additional reachability cost function. Furthermore, to estimate the displacement of the target object given an action trajectory, we implement a stochastic dynamics model $\configdynest$. 
We use the cost functions and the dynamics function inside MPC, which executes in a closed loop for $\nsteps$ steps. During this process, we detail how to merge current observations with previous ones and update the pose posterior $p(\T | \xall)$.

\subsection{Representing Pose Posterior}\label{RUMI:sec:posterior}
We maintain a belief over the pose posterior $p(\T|\xallt)$ using a particle filter, where each particle is a pose. 
We have $\numparticles$ particles $\T_{1..\numparticles}$, with weights $\weight_{1..\numparticles}$ such that $\sum_{i=1}^\numparticles \weight_i = 1$.
Our choice of a particle filter over alternative representations is motivated by the potential multi-modality of the posterior and the ability to process each particle in parallel.

A major obstacle to the tractability of solving Eq.~\ref{RUMI:eq:mi objective} is the information correlation between \descriptor{}s. Observing one decreases the information gain from others in a non-trivial manner, and it is a common long-standing assumption to consider the information gain from each independently~\cite{cao2013multi},~\cite{stachniss2003exploring}. Thus, we assume the conditional mutual independence of
$\Sx$ for all query positions $\x$ given observed $\xallt$.

Critical to our method is a way to evaluate the 
posterior $p(\T | \xall)$.
Our prior work CHSEL~\cite{zhong2023chsel} formulated a differentiable cost function $\relaxedtotalcost(\xall, \T)$ that evaluates the discrepancy between $\xall$ and $\T$. It bears similarity to hydroelastic, or pressure field contact modelling~\cite{elandt2019pressure},~\cite{masterjohn2022velocity}, except in addition to the pressure field penalizing object penetration, there are pressure fields that penalize semantics violation, such as observed free space \descriptor{}s being inside objects.

We simplify the third semantics class from CHSEL, which represented known SDF of any value. We restrict it to $s=0$, which refers to surface points. The cost is formulated by first partitioning the observed $\xall$ into $\xfree = \{(\x,\s) \ | \ \s = \free \}$, $\xocc = \{(\x,\s) \ | \ \s = \occupied\}$, and $\xsurface = \{(\x,\s) \ | \ \s = \surface \}$. 
\begin{align}\label{RUMI:eq:cost}
    \relaxedtotalcost(\xall, \T) &= \sum_{\mathclap{\x,\s \in \xfree}}  \freecost(\xtrans) + \sum_{\mathclap{\x,\s \in \xocc}}  \occupiedcost(\xtrans) + \sum_{\mathclap{\x,\s \in \xsurface}}  \knowncost(\xtrans)\\
        \freecost(\xtrans) &= \scalefree \max(0, \interiorthreshold-\sdf(\xtrans)) \\
        \occupiedcost(\xtrans) &= \scalefree \max(0, \interiorthreshold + \sdf(\xtrans))\\
        \knowncost(\xtrans) &= |\sdf(\xtrans)|
\end{align}
where $\scalefree > 0$ is a scaling parameter and $\interiorthreshold > 0$ allows for small degrees of violation due to uncertainty in the positions.

Their gradients are defined as
\begin{align}\label{RUMI:eq:grad}
    \nabla \freecost(\xtrans) &= \scalefree \max(0, \interiorthreshold-\sdf(\xtrans)) (-\nabla \sdf(\xtrans))\\
    \nabla \occupiedcost(\xtrans) &= \scalefree \max(0, \interiorthreshold + \sdf(\xtrans)) \nabla \sdf(\xtrans)\\
    \nabla \knowncost(\xtrans) &= \sdf(\xtrans) \nabla \sdf(\xtrans)
\end{align}
where $\nabla \sdf(\xtrans)$ is the object SDF gradient with respect to an object-frame position $\xtrans$ and normalized such that $\norm{\nabla \sdf(\xtrans)}_2 = 1$.

Similar to energy-based methods, we use the Boltzmann distribution~\cite{haarnoja2017reinforcement},~\cite{teh2003energy} to interpret Eq.~\ref{RUMI:eq:cost} as the posterior pose probability:
\begin{equation}\label{RUMI:eq:posterior}
    p(\T | \xall) = \pnormalization e^{-\peakiness \relaxedtotalcost(\xall, \solution)}
\end{equation}
where $\peakiness > 0$ selects how peaky the distribution should be and $\pnormalization$ is the normalization constant such that 
$\int \pnormalization e^{-\peakiness \relaxedtotalcost(\xall, \solution)} d\T = 1$.

We observe that the cost in Eq.~\ref{RUMI:eq:cost} is additive in the sense 
\begin{equation}\label{RUMI:eq:additive}
\relaxedtotalcost(\xall \cup (\x,s), \T) = \relaxedtotalcost(\xall, \T) + \relaxedtotalcost((\x,s), \T)
\end{equation}
This is an important property that enables us to efficiently evaluate information gain of all workspace positions in parallel.


\subsection{Mutual Information Surrogate}
Our conditional mutual independence assumption of $p(\Sx|\xall)$ lets us consider the information gain from knowing the semantics at a single new position, which we denote the \textit{\infofieldname{}} $\infofield(\x|\xall)$. This is much simpler than considering the information gain of a robot trajectory directly because there is no time component or correlation between the semantics of neighbouring \descriptor{}s. Suppose we have observed $\xall$ and want to evaluate the information gain from observing some new \descriptor{} $(\x, \s)$.
Note that here we are querying a specific given value of $\x$, but $\Sx$ is still a random variable, so the expectation is over
$p(\Sx|\xall)$:

\begin{align}
    \infoklfor(\x|\xall) &= \Expect_{\s \sim p(\Sx|\xall)} [D_{KL} (  p(\solution | \xall \cup (\x,\s)) || p(\solution|\xall))]\\
    &= \Expect_{\s} [\Expect_{\solution \sim p(\solution | \xall \cup (\x, \s))} [\log \frac{p(\solution | \xall \cup (\x, \s))}{\pprior}] ]
\end{align}

The forward KL divergence results in an expectation over $p(\solution | \xall \cup (\x, \s))$. Since we need to evaluate the information gain for many positions in the workspace, this becomes intractable.

To address this challenge, we use the reverse KL divergence, 
since the expectation is then over $p(\solution|\xall)$ for all queried positions. 
In general, KL divergence is not symmetric. However, when two distributions are close together the KL divergence is approximately symmetric~\cite{zhang2024properties},~\cite{kullback1997information}. In our case the KL divergence is between $\pprior$ and $\pposterior$ with all having $\poseset$ support, avoiding infinite divergences. As we increase $|\xall|$ during exploration, we expect the two distributions to become closer and the reverse KL to better approximate the forward KL divergence.

Intuitively, a \descriptor{} has high reverse KL divergence if it has high $\pprior$ and low $\pposterior$. These correspond to \descriptor{}s that would invalidate currently high-probability poses i.e. these are positions we would like to explore.

Using reverse KL, We now have
\begin{align}\label{RUMI:eq:revkl}
    \infoklrev(\x|\xall) &= \Expect_\s [D_{KL} ( p(\solution|\xall) || p(\solution | \xallplus))]\\
    &= \Expect_{\s} [\Expect_{\solution \sim \pprior} [\log \frac{\pprior}{p(\solution | \xallplus)}] ]
\end{align}

Substituting Eq.~\ref{RUMI:eq:posterior} in
\begin{align}
    \infoklrev(\x|\xall) &= \Expect_{\s} [\Expect_{\T} [\log \frac{p(\T|\xall)}{p(\T | \xall \cup (\x, \s) )}]]\\
    &= \Expect_{\s} [\Expect_{\solution}[\log \frac{\pnormalization_1 e^{-\peakiness\relaxedtotalcost(\xall, \solution)}}{\pnormalization_2 e^{-\peakiness\relaxedtotalcost(\xall  \cup (\x, \s), \solution)}}]]\\
    &= \Expect_{\s} [\Expect_{\solution}[\log \frac{e^{-\peakiness \relaxedtotalcost(\xall, \solution)}}{e^{-\peakiness \relaxedtotalcost(\xall  \cup (\x, \s), \solution)}}]] + \log \frac{\pnormalization_1}{\pnormalization_2}\\
    &= \peakiness \Expect_{\s} [\Expect_{\solution}[-\relaxedtotalcost(\xall, \solution) + \relaxedtotalcost(\xall  \cup (\x, \s), \solution))] + \log \frac{\pnormalization_1}{\pnormalization_2}
\end{align}

\noindent where $\pnormalization_1$ and $\pnormalization_2$ are the normalizing constants for $p(\T|\xall)$ and $p(\T|\xallplus)$, respectively. First we simplify using the additive property of $\relaxedtotalcost$ (Eq.~\ref{RUMI:eq:additive}) then consider the normalizing constants,

\begin{align}\label{RUMI:eq:revkl normalizing constant}
    \infoklrev(\x|\xall) &= \peakiness \Expect_{\s} [\Expect_{\T} [\relaxedtotalcost((\x,\s),\solution)]]+ \log \frac{\pnormalization_1}{\pnormalization_2}
\end{align}

We note that $\pnormalization_2$ depends on the querying position $\x$ because each $\x$ induces a different $p(\T|\xallplus)$.
This normalizing constant is intractable to compute because it involves an integral over $\T$, so we instead optimize the approximation
\begin{align}
    \infofield(\x | \xall) &= \peakiness \Expect_{\s} [\Expect_{\solution}[\relaxedtotalcost((\x,\s),\solution)]] \\
    &= \peakiness \sum_{\s} p(\Sx = \s | \xall ) \Expect_{\solution}[ \relaxedtotalcost((\x,s),\solution)]
\end{align}
Selecting $\peakiness$ too high leads to the pose particle weights dominated by a few, causing particle degeneracy.

We approximate the expectation over the posterior by taking the weighted sum over the pose particles
\begin{equation}
    \infofield(\x|\xall) \approx \peakiness \sum_{\s} \sum_{i=1}^{\numparticles} p(\Sx = \s | \xall ) \weight_i \relaxedtotalcost((\x,s),\T_i)\label{RUMI:eq:info}
\end{equation}

\begin{figure}
    \centering
    \includegraphics[width=0.32\linewidth]{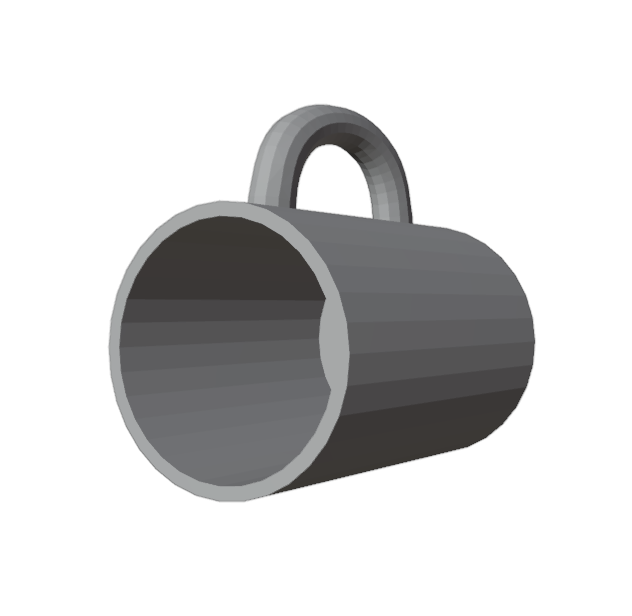}
    \includegraphics[width=0.32\linewidth]{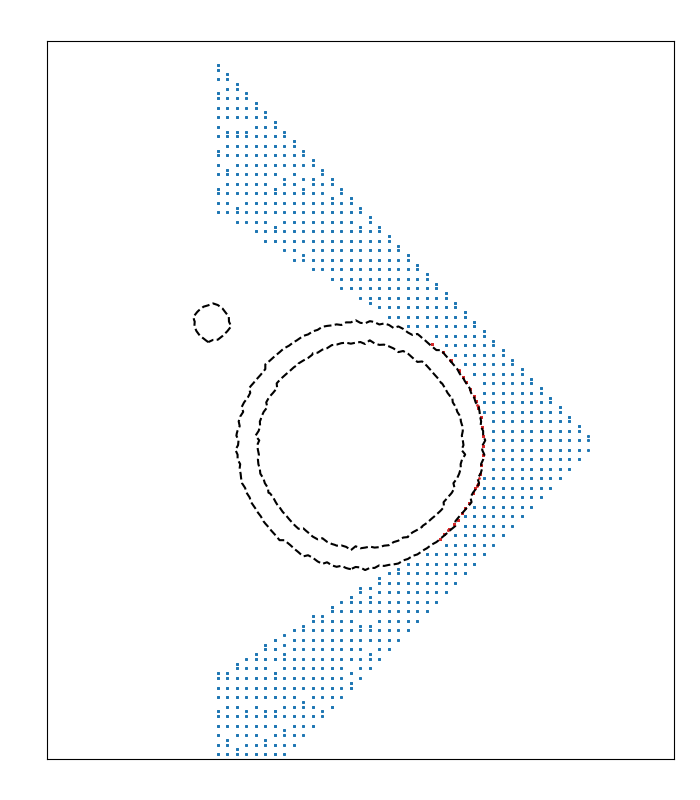}
    \includegraphics[width=0.32\linewidth]{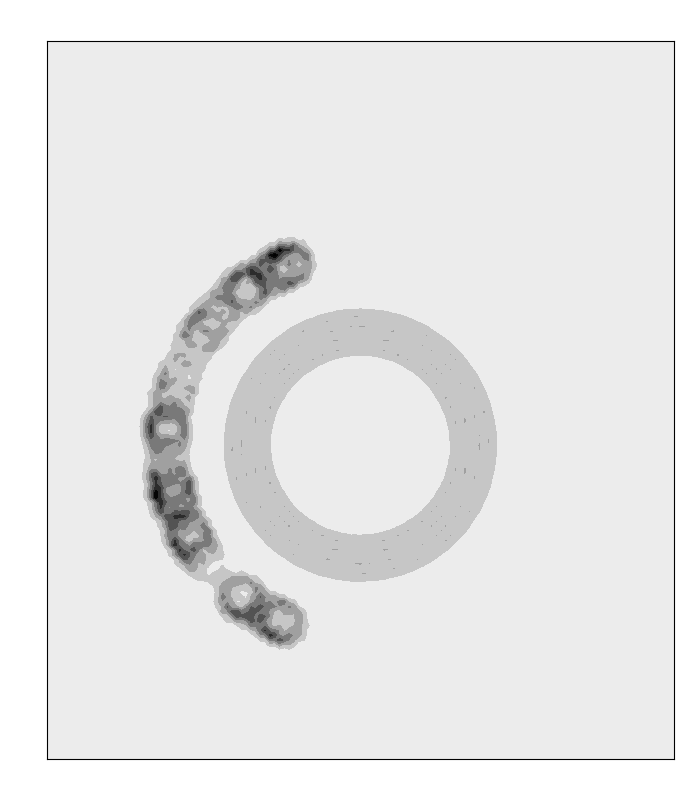}
    \caption{(left) Example mug object with (middle) $\xall$ rendered from a pinhole camera on one side, not seeing where the handle is. The ground truth object surface is outlined in dotted black, observed surface \descriptor{}s are in red, and observed free space is in blue. (right) The \infofieldname{} estimated with $\numparticles=100$ is darker where there is more information, where the handles could be.}
    \label{RUMI:fig:intuition}
\end{figure}


Finally, we consider how we can approximate the conditional semantics distribution $p( \Sx | \xall)$ which is the last term required for fully computing $\infofield(\x | \xall)$. We use the law of total probability
\begin{align}
    p( \Sx | \xall) &= \int  p(\Sx | \solution, \xall) p(\T | \xall) d\T
\end{align}
Here again we approximate the expectation over the posterior by taking the weighted sum over the pose particles
\begin{align}
    p(\Sx|\xall) &\approx \sum_{i=1}^{\numparticles} \weight_i p(\Sx | \T_i, \xall)\label{RUMI:eq:approxsem}
\end{align}
we assume the conditional independence of $\Sx$ and $\xall$ when given $\T$, so
\begin{align}
    p(\Sx|\xall) &\approx  \sum_{i=1}^{\numparticles}  \weight_i p(\Sx | \T_i))\\
     &=  \sum_{i=1}^{\numparticles}  \weight_i p(\Sx | \sdf(\T_i \x))\label{RUMI:eq:calcsem}
\end{align}
where $p(\Sx | \sdf(\T_i \x))$ is given by the sensor model.

Note that all the terms in Eq.~\ref{RUMI:eq:info} only query $\x$ and $\T_{1..\numparticles}$, without needing to directly consider $\xall \cup (\x,\s)$. This enables us to evaluate $\infofield(\x | \xall)$ for all positions inside a workspace $\x \in \workspace \subset \reals^3$ in parallel. 

\subsection{Illustrative Example}
To develop intuition, we consider a mug as the target object, depicted in Fig.~\ref{RUMI:fig:intuition} (left). Initially, a camera observes one side of the mug, narrowing down its position. However, since it cannot observe the handle and there is partial rotational symmetry, there is uncertainty in the orientation of the object. Fig.~\ref{RUMI:fig:intuition} (right) is the computed information gain $\infofield(\x|\xall)$ over
the entire workspace, showing that most of the information
gain is concentrated where the handle could be. 

\begin{figure}
    \centering
    \includegraphics[width=0.32\linewidth]{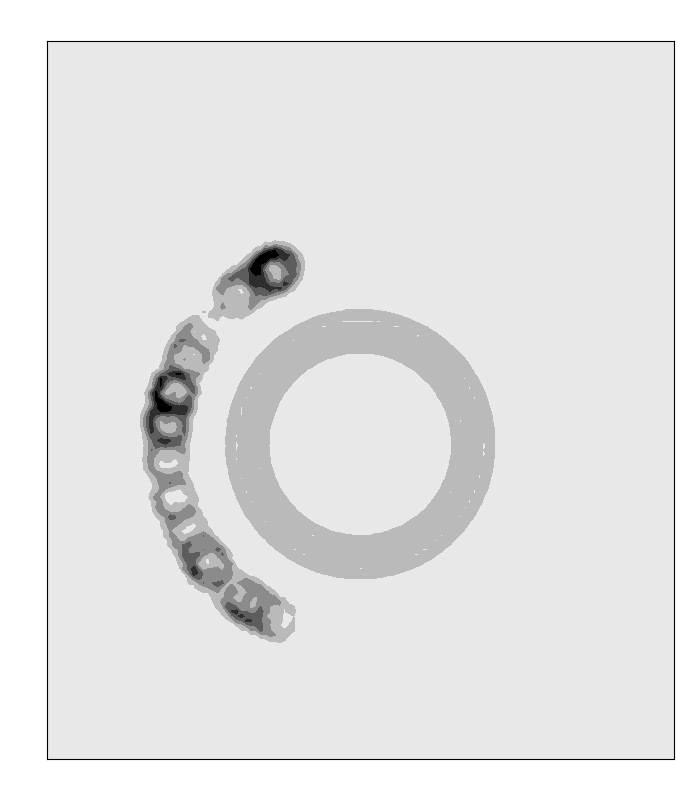}
    \includegraphics[width=0.32\linewidth]{media/intuition/9_lambda_2_alpha_100_just_kl.png}
    \includegraphics[width=0.32\linewidth]{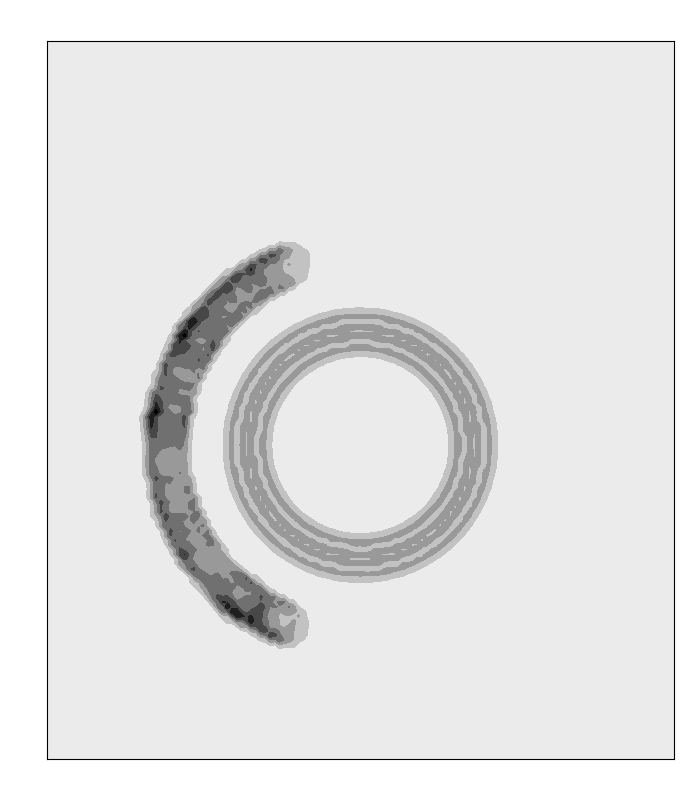}
    \caption{Computed \infofieldname{} $\infofield(\x | \xall)$ from Fig.~\ref{RUMI:fig:intuition} with (left) $\numparticles=30$, (middle) $\numparticles=100$, and (right) $\numparticles=1000$.}
    \label{RUMI:fig:numparticles}
\end{figure}

Intuitively, we expect a smooth dark band where the handles could be but observe unevenness. This is due to the approximation error of $\pprior$ being represented by finitely many pose particles and the approximation of $\infoklrev$ with $\infofield$. This is illustrated by Fig.~\ref{RUMI:fig:numparticles}. With larger $\numparticles$, the trade off for gaining a more accurate approximation of $\pprior$ is increased memory usage. Since we process the particles and query positions in parallel, memory becomes the bottleneck and they have to be processed in batches, turning the memory trade off into a runtime one.

\begin{figure*}
    \centering
    \includegraphics[width=0.32\linewidth]{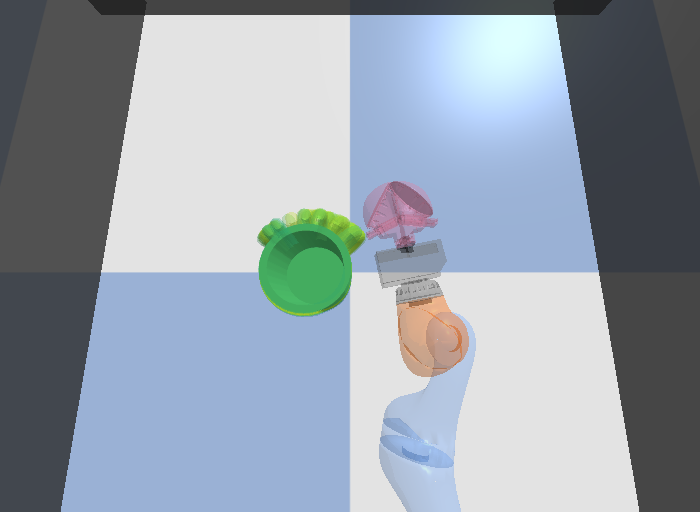}
    \includegraphics[width=0.32\linewidth]{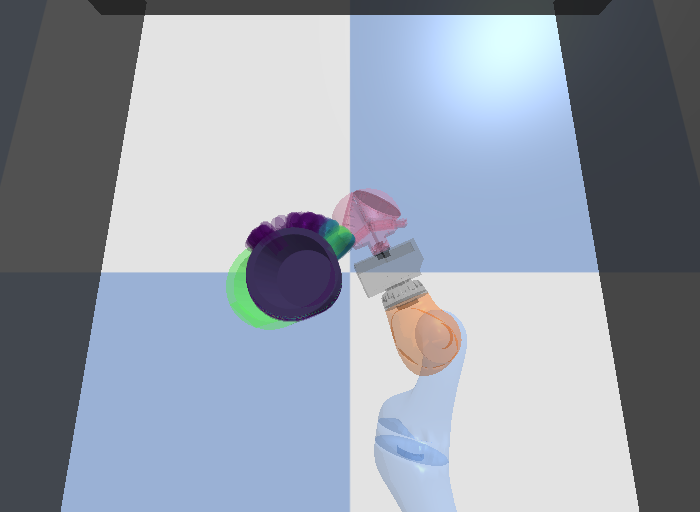}
    \includegraphics[width=0.32\linewidth]{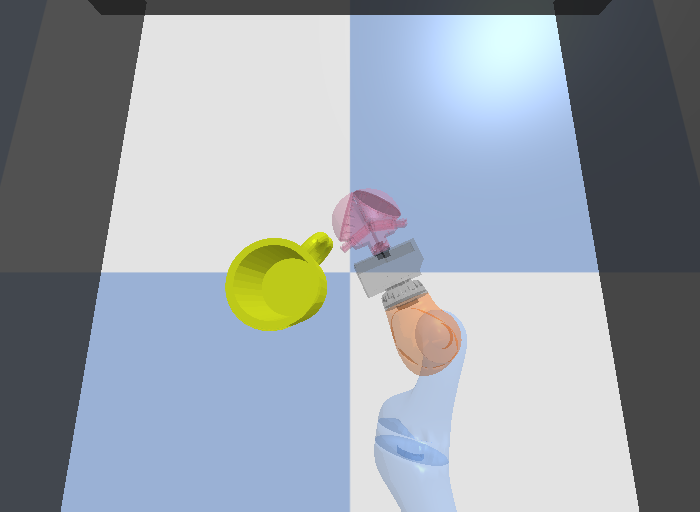}
    \caption{Resampling during a pybullet simulated mug task with partial initial observation like in Fig.~\ref{RUMI:fig:intuition}. (left) Before contact pose particles, and (middle) pose particles after contact with many receiving low likelihood indicated by the dark color due to discrepancy with the newly observed surface \descriptor{}s. This leads to resampling, and (right) resampled particles that are all high likelihood and centered around the ground truth pose.}
    \label{RUMI:fig:resample}
\end{figure*}

\newcommand{\dt}{\boldsymbol{\Delta t}}
\newcommand{\dr}{\boldsymbol{\Delta R}}

\begin{algorithm}[t]
\DontPrintSemicolon
\SetKwInput{Given}{Given}
\SetKwInput{Hyperparameters}{Hyperparameters}

\Given{
$\numparticles$ number of particles,\\
$\Tprior$ initial poses,\\
$\config_1$ initial robot configuration,\\
$\updatenoiset$ translation noise,\\
$\updatenoiser$ rotation noise,\\
$\resamplethreshold$ resample discrepancy threshold,\\
$\workspace$ workspace set of query positions\\
}
$\xall_0 \leftarrow \text{sensors observe at } \config_1$ \;
$\T_{1..\numparticles} \leftarrow \chsel(\Tprior, \xall_0)$\;\label{RUMI:line:init}
$\weight_{1..\numparticles} \leftarrow \pweight(\T_{1..\numparticles}, \xall_0)$ \;
\For{$t \leftarrow 1$ \KwTo $\nsteps$}{
    compute $p(\Sx|\xall_{t-1})$ using Eq.~\ref{RUMI:eq:calcsem} and $\infofield(\x|\xall_{t-1})$ using Eq.~\ref{RUMI:eq:info} for $\x \in \workspace$ and cache in voxel grids \;
    $\us_t \leftarrow $ \texttt{Plan}($\infofield(\x|\xall_{t-1}), p(\Sx|\xall_{t-1}), \config_t$)\label{RUMI:line:execute}\;
    robot executes action $\us_t$ to arrive at $\config_{t+1}$\;
    $\xthis_t \leftarrow $ sensors observe at $\config_{t+1}$\label{RUMI:line:observe}\;
    $\Delta \T_r \leftarrow$ sensors observe change in robot end-effector pose while in contact\label{RUMI:line:objmove}\;
    $\Delta \T, \Delta \T_w \leftarrow \pxestimatemove(\xall_{t-1},\xthis_t, \T_{1..\numparticles}, \Delta \T_r)\label{RUMI:line:estimatemove}$\;
    \uIf{$\Delta \T$ not $\mathbf{0}$} {
        \tcp{predict step}
        \For{$i \leftarrow 1$ \KwTo $\numparticles$}{
            $\Delta \T_\sigma \leftarrow \pperturb{}(\updatenoiset, \updatenoiser)$
            $\T_i \leftarrow \Delta \T_\sigma  \cdot \Delta \T \cdot \T_i  $\;
        }
    }
    $\xallt \leftarrow \pxmergeobs(\xall_{t-1},\xthis_t,\T_{1..\numparticles}, \Delta \T_w)$\label{RUMI:line:mergeobs}\;
    \tcp{update step, even when not in contact}
    $\discrepancy_{1..\numparticles} \leftarrow \relaxedtotalcost(\xall, \T_{1..\numparticles}$)\;
    \uIf{$\max(\discrepancy_{1..\numparticles}) > \resamplethreshold$}{\label{RUMI:line:doresample}
        $\T_{1..\numparticles} \leftarrow \resample(\T_{1..\numparticles}, \weight_{1..\numparticles}, \xall_t)$\label{RUMI:line:resample}\;
        $\weight_{1..\numparticles} \leftarrow 1 / \numparticles$\; 
    }
    \uElse{
    $\weight_{1..\numparticles} \leftarrow \pweight(\T_{1..\numparticles}, \xall_t )$
    }
}
\caption{Particle filter posterior update}
\label{RUMI:alg:update}
\end{algorithm}

\begin{algorithm}
\DontPrintSemicolon
\SetKwInput{Given}{Given}
\SetKwInput{Hyperparameters}{Hyperparameters}
\SetKwInOut{Output}{Output}

\Given{
$\T_{1..\numparticles}$ pose particles,\\
$\xall$ set of observed \descriptor{}s,\\
$\peakiness$ peakiness
}
$\discrepancy_{1..\numparticles} \leftarrow \relaxedtotalcost(\xall, \T_{1..\numparticles}$)\;
$\weight_{1..\numparticles} \leftarrow e^{-\peakiness \discrepancy_{1..\numparticles}}$ \label{RUMI:line:weight}\;
$\weight_{1..\numparticles} \leftarrow \weight_{1..\numparticles} / \sum_{i=1}^\numparticles \weight_i$ \tcp{normalize sum to 1}
\caption{\pweight{}}
\label{RUMI:alg:weigh}
\end{algorithm}

\subsection{Posterior Update Process}\label{RUMI:sec:update}
So far, we have developed the \infofieldname{} given some observed $\xall$ at one time step. We now describe the active rummaging process in Algorithm~\ref{RUMI:alg:update} to update the posterior.

Before any actions, we are given the pose prior $p(\T)$ in the form of $\numparticles$ initial poses $\Tprior$. Note that given a fixed set of \descriptor{}s $\xall$, the 
posterior probability
of poses can be compared using Eq.~\ref{RUMI:eq:posterior}. With the relative 
posterior probability
and samples from the prior, we can theoretically draw samples from the posterior using techniques such as Markov Chain Monte Carlo (MCMC)~\cite{geyer1992practical},~\cite{casella1992explaining}. However, MCMC tend to struggle with the high dimensionality of poses ($\T \in \poseset)$. With the interpretation of Eq.~\ref{RUMI:eq:cost} as the 
posterior
(Eq.~\ref{RUMI:eq:posterior}), optimization of Eq.~\ref{RUMI:eq:cost} on prior pose particles can naturally be interpreted as approximately sampling from the posterior. Thus we apply CHSEL (Algorithm 1 from~\cite{zhong2023chsel}) to produce the initial pose particles in Algorithm~\ref{RUMI:alg:update} line~\ref{RUMI:line:init}. 
CHSEL performs Quality Diversity (QD) optimization~\cite{pugh2016quality} on Eq.~\ref{RUMI:eq:cost} to find poses that have low discrepancy while maintaining diversity across some measure of pose space. We use the orientation component of $\T$, or just the yaw when restricting the pose search space to $\posesetplanar$ as the measure.

We then assign weights to each pose particle as described in Algorithm~\ref{RUMI:alg:weigh}. These weights represent the relative posterior probability of each particle. We normalize the weights so that $\sum_{i=1}^{\numparticles} \weight_i = 1$. Normalizing is important so that the use of weights in approximating expectations over the pose posterior in Eq.~\ref{RUMI:eq:info} and Eq.~\ref{RUMI:eq:approxsem} remain valid. A side benefit of normalization is that we can omit the normalizing constant $\pnormalization$ from Eq.~\ref{RUMI:eq:posterior} in Algorithm~\ref{RUMI:alg:weigh} line~\ref{RUMI:line:weight}. 

Then for each time step $t$, we first compute $p(\Sx|\xall_{t-1})$ and $\infofield(\x|\xall_{t-1})$ using Eq.~\ref{RUMI:eq:calcsem} and Eq.~\ref{RUMI:eq:info}, respectively, for $\x \in \workspace$ and cache the results in voxel grids. These voxel grids allow linear interpolation querying and return 0 for $\infofield(\x|\xall_{t-1})$ and $\free$ for $p(\Sx|\xall_{t-1})$ when 
$\x$ is outside $\workspace$. 
They are used to plan a robot trajectory, as described in Subsection~\ref{RUMI:sec:planning}. The robot executes the first action in the planned trajectory and sensors observe both a new set of \descriptor{}s $\xthis_t$ and the change in robot end effector pose while in contact $\Delta \T_r$.

In Algorithm~\ref{RUMI:alg:update} line~\ref{RUMI:line:estimatemove} we estimate the change in pose $\Delta \T$ of the target object given $\xall_{t-1}$, $\xthis_t$, $\T_{1..\numparticles}$, and $\Delta \T_r$. Some end-effectors can either enforce sticking contact~\cite{jaiswal2017vacuum} or measure slip (such as in~\cite{melchiorri2000slip},~\cite{romeo2020methods}) to estimate $\Delta \T$ directly. Not all robots have these sensors, so we present an optimization based method in Algorithm~\ref{RUMI:alg:objmove}. The main idea is to find a $\Delta \T$ that transforms $\xall_{t-1}$ such that it is consistent with the most recently observed $\xthis_t$. Our prior is that contact was sticking; that is $\Delta \T = \Delta \T_r$ in Algorithm~\ref{RUMI:alg:objmove} line~\ref{RUMI:line:sticky}. We select a representative pose particle $\T_i$ with the lowest discrepancy to apply $\Delta \T$ to.
For $\noptim$ optimization steps, we evaluate $\relaxedtotalcost$ on $\xthis$ and the hypothesis new pose $\Delta \T \cdot \T_i$. $\relaxedtotalcost$ is differentiable with respect to $\Delta \T \cdot \T_i$, and we back propagate gradients to $\Delta \T$ and perform stochastic gradient descent (SGD). We then produce the world frame change in pose $\Delta \T_w = \T_i^{-1} \cdot \Delta \T \cdot \T_i$ that can be applied to world frame positions.

Typically in particle filters we update the posterior via alternating prediction (via forward dynamics) and correction (from sensor data)
steps. If the object did not move, then we also predict the pose particles remain stationary. If the object did move ($\Delta \T$ not $\mathbf{0}$) then our forward dynamics predicts movement   
$\T_i \leftarrow \Delta \T_\sigma  \cdot \Delta \T \cdot \T_i$,
where $\Delta \T_\sigma$ is a transform perturbation sampled with the process in Algorithm~\ref{RUMI:alg:perturb} that adds diversity to the particles.

Before we can perform the correction step, we first merge the previous observations $\xall_{t-1}$ with the current observations $\xthis_t$, as described in Algorithm~\ref{RUMI:alg:mergeobs}. The object \descriptor{}s are transformed by $\Delta \T_w$ while the free \descriptor{}s remain stationary. However, the move might have invalidated some previous free ones and so we check whether $\sdf(\T_i \x) > 0, \ \forall i = 1..\numparticles$ for each $(\x,\free) \in \xall$ in Algorithm~\ref{RUMI:alg:mergeobs} line~\ref{RUMI:line:freeconsistency}. We then take the union of the transformed $\xall_o$, validated $\xfree'$, and newly observed $\xthis$. To avoid duplicate data, we voxel downsample by creating voxel grids, one per semantics value, that spans the range of the positions with resolution $\resdownsample$. We assign the voxel grids with the positions then extract the center of voxel cells that received any assignment as new positions. We denote this downsampling process as $D: \reals^3 \times \reals^+ \rightarrow \reals^3$.

With updated observations $\xall_t$, we can update the weights of the pose particles. Importantly, we update even when not making contact because observing $\s = \free$ \descriptor{}s provides information about where the object is not. This process is described in Algorithm~\ref{RUMI:alg:weigh}, where Eq.~\ref{RUMI:eq:cost} is applied to get discrepancies $\discrepancy_{1..\numparticles}$. We then apply Eq.~\ref{RUMI:eq:posterior} to convert it to an unnormalized 
probability.
For numerical stability, we subtract the minimum $\discrepancy$ from all of them to get relative discrepancy. This is without loss of generality since the normalization forces the weights to sum to 1.

In addition to the update step, we resample the pose particles to avoid degeneracy and maintain diversity as is typical of particle filters. Many heuristics exist for deciding when to resample~\cite{li2015resampling} based mostly on removing low weight particles. However, the particle weights only represent their relative probability with respect to other particles, and we have a more direct signal in the discrepancy $\discrepancy_{1..\numparticles} = \relaxedtotalcost(\xall, \T_{1..\numparticles})$ to evaluate when the pose particles have low likelihood. We use this in Algorithm~\ref{RUMI:alg:update} line~\ref{RUMI:line:doresample} by comparing the maximum discrepancy of the particles to a threshold $\resamplethreshold > 0$. For better robustness against outlier pose samples, a percentile of the discrepancy instead of the max can be used. This process is visualized in Fig.~\ref{RUMI:fig:resample}, where a contact made with the handle at the back of the mug forces a resample due to the previous pose particles' discrepancy with the observed surface \descriptor{}s.

Finally, the resampling process is described in Algorithm~\ref{RUMI:alg:resample}. We first perform the well known sampling importance resampling~\cite{li2015resampling}, then like in the prediction step we perturb the pose particles to generate diversity. We then ensure the pose particles have high probability by performing SGD on $\relaxedtotalcost(\xall, \T_{1..\numparticles})$. 

\begin{algorithm}[t]
\DontPrintSemicolon
\SetKwInput{Given}{Given}
\SetKwInput{Hyperparameters}{Hyperparameters}
\SetKwInOut{Output}{Output}

\Given{
$\updatenoiset$ translation noise, $\updatenoiser$ rotation noise
}
\Output{
$\Delta \T_\sigma$ delta transformation
}
    \tcp{sample process noise}
    $\dt \sim \normal(\zero, \diag([\updatenoiset, \updatenoiset, \updatenoiset])$\;
    $\theta \sim \normal(0, \updatenoiser)$\;
    $\mathbf{e} \sim U(\{\x \ | \ \norm{\x}_2 = 1, \x \in \reals^3\})$\label{RUMI:line:axissample}\;
    $\dr \leftarrow e^{\theta \mathbf{e}}$\tcp{axis angle to matrix}
    $\Delta \T_\sigma \leftarrow \begin{bmatrix}\dt & \dr\\ \mathbf{0} & 1\end{bmatrix}$\;
\caption{\pperturb{}} 
\label{RUMI:alg:perturb}
\end{algorithm}

\begin{algorithm}[t]
\DontPrintSemicolon
\SetKwInput{Given}{Given}
\SetKwInput{Hyperparameters}{Hyperparameters}

\Given{
$\T_{1..\numparticles}$ pose particles,\\
$\weight_{1..\numparticles}$ particle weights,\\
$\xall$ set of observed \descriptor{}s,\\
$\updatenoiset$ translation noise,\\
$\updatenoiser$ rotation noise\\
$\noptim$ resample optimization steps
}
\tcp{sampling importance resampling}
$\T_{1..\numparticles} \leftarrow \sir(\T_{1..\numparticles}, \weight_{1..\numparticles})$\;
\For{$i \leftarrow 1$ \KwTo $\numparticles$}{
    $\Delta \T_\sigma \leftarrow \pperturb{}(\updatenoiset, \updatenoiser)$
    $\T_i \leftarrow \Delta \T_\sigma  \cdot \T_i$\;
}
\For{$j \leftarrow 1$ \KwTo $\noptim$} {
    differentiate $\relaxedtotalcost(\xall, \T_{1..\numparticles})$ to get $\T_\onetoparticles$ gradients \;
    SGD to optimize $\T_\onetoparticles$\;
}
\caption{\resample{}}
\label{RUMI:alg:resample}
\end{algorithm}

\begin{algorithm}[t]
\DontPrintSemicolon
\SetKwInput{Given}{Given}
\SetKwInput{Hyperparameters}{Hyperparameters}

\Given{
$\xall$ set of previously observed \descriptor{}s\\
$\xthis$ set of new observed \descriptor{}s\\
$\T_{1..\numparticles}$ pose particles,\\
$\Delta \T_r$ change in end-effector pose during contact\\
$\noptim$ number of optimization steps
}
$\Delta \T \leftarrow \Delta \T_r$ \tcp{sticking contact prior}\label{RUMI:line:sticky}
$\discrepancy_{1..\numparticles} \leftarrow \relaxedtotalcost(\xall, \T_{1..\numparticles}$)\;
$i \leftarrow \argmin \discrepancy_{1..\numparticles}$\;
\For{$j \leftarrow 1$ \KwTo $\noptim$}{
    \tcp{hypothesis moved object pose}
    differentiate $\relaxedtotalcost(\xthis, \Delta \T \cdot \T_i )$ to get $\Delta \T$ gradients \;
    SGD to optimize $\Delta \T$\;
}
$\Delta \T_w \leftarrow \T_i^{-1} \cdot \Delta \T \cdot \T_i$
\caption{\pxestimatemove{}} 
\label{RUMI:alg:objmove}
\end{algorithm}

\begin{algorithm}
\DontPrintSemicolon
\SetKwInput{Given}{Given}
\SetKwInput{Hyperparameters}{Hyperparameters}

\Given{
$\xall$ set of previously observed \descriptor{}s,\\
$\xthis$ set of new observed \descriptor{}s,\\
$\T_{1..\numparticles}$ pose particles,\\
$\Delta \T_w$ world frame change in object pose\\
$\resdownsample$ downsample resolution
}
$\xall_o \leftarrow \{(\Delta \T_w \x,\s) \ | \ (\x,\s) \in \xall, \s \neq \free \}$ \;
$\xfree \leftarrow \{(\x,\s) \ |  \ (\x,\s) \in \xall, \s = \free \}$ \tcp{stationary}
\tcp{remove all that may be occupied now}
$\xfree' \leftarrow \{ (\x,\s) \ |  \ (\x,\s) \in \xfree, \sdf(\T_i \x) > 0, \ \forall i = 1..\numparticles\}$\label{RUMI:line:freeconsistency}\;
$\xall \leftarrow \xall_o \cup \xfree' \cup \xthis$\;
voxel downsample $\xall$ with resolution $\resdownsample$
\caption{\pxmergeobs{}}
\label{RUMI:alg:mergeobs}
\end{algorithm}


\subsection{Planning Problem}\label{RUMI:sec:planning}

We use model predictive path integral (MPPI) control~\cite{williams2017model} to plan a $\nhorizon$ horizon length trajectory and execute the first step of it in Algorithm~\ref{RUMI:alg:update} line~\ref{RUMI:line:execute}. $\nhorizon$ may be less than $\nsteps$ due to computation limitations. Without loss of generality, consider $t=1$ at planning time for notation simplification. MPPI samples many Gaussian action perturbations around a nominal action trajectory to produce $\us_{1..\nhorizon}$, rolls out the robot configuration from $\config_0$ to get $\config_{1..\nhorizon}$ with a dynamics function, and evaluates each configuration trajectory with a cost function to weigh how the action trajectories should be combined. We initialize the nominal trajectory with noise, warm start it by running MPPI without actually executing the planned trajectory for several iterations. Then when executing $\us_{1}$, we use $\us_{2..H},\mathbf{0}$ as the nominal trajectory for the next step. 
By convention, MPPI minimizes cost, and so we present costs where lower values are better.

\subsection{Information Gain Cost}
We assume we have the robot model such that we can map $\interiorobs(\config) \rightarrow \interiorset$ where $\interiorset$ is the set of world coordinate positions inside or on the surface of the robot. Note that when observing $\xthis_t$ in Algorithm~\ref{RUMI:alg:update} line~\ref{RUMI:line:observe}, $\{(\x,\free) \ | \ \x \in \interiorobs(\config_{t+1})\}$ should at least be in $\xthis_t$ since the object cannot be inside the robot.
Additionally, we assume we can identify $\interiorobsinfo(\config) \subset \interiorobs(\config)$ that selects the points of the robot that can observe information through contact. For example, the wrist of the end effector may be much less effective at reliably localizing contact than the tactile sensor. We only consider $\interiorobsinfo$ for gathering information but the full $\interiorobs$ for the dynamics model. 

For a rolled-out configuration trajectory $\config_{1..\nhorizon}$ we define the information gain cost
\begin{equation}\label{RUMI:eq:costinfo nomove}
    \costinfo'(\config_{1..\nhorizon}) = \sum_{\x \in \downsample(\bigcup _{i=1}^\nhorizon \interiorobsinfo(\config_i),\resdownsample)} - \infofield(\x | \xall)
\end{equation}
which is the \infofieldname{} at every robot interior point in the rolled out trajectory, downsampled to avoid double-counting.

This cost function develops naturally from $\infofield(\x | \xall)$, however it does not take into account that the object can move, and in doing so, can change $\infofield(\x | \xall)$. Consider a trajectory where a robot moves into contact with the target object then continues in a straight line with the target object remaining in sticking contact. While it would traverse the workspace and gather high $\costinfo'$ as a result, relative to the object it has not moved after coming into contact, and so should collect no new information. Indeed, $\infofield(\x | \xall)$ is better seen as an object frame field, as only motion relative to the object should collect information.

To address this, we introduce predicted object displacement $\displacement \in \reals^3$, and define the adjusted information gain cost
\begin{equation}\label{RUMI:eq:costinfo}
    \costinfo(\config_{1..\nhorizon}, \displacement_{1..\nhorizon}) = \sum_{\x \in \downsample(\bigcup _{i=1}^\nhorizon [\interiorobsinfo(\config_{i}) - \displacement_i],\resdownsample) } - \infofield(\x | \xall)
\end{equation}
where $\interiorobs(\config_i) - \displacement_i \ \forall i=1..\nhorizon$ transforms the world query positions to be in the displaced object frame.

\subsection{Dynamics Model}
We predict the displacement $\displacement$ in our dynamics model $\configdynest$ in addition to $\config$. We assume the difference of the true dynamics $\configdyn$ from the given free space dynamics $\configdyngiven$ is only due to making contact with the target object, and use the precomputed $p(\Sx|\xall)$ voxel grid to predict when that occurs. One step of $\configdynest$ is described in Algorithm~\ref{RUMI:alg:dynamics} and below: 

\newcommand{\surfnorm}{\hat{\textbf{n}}}
First we apply free space dynamics to get candidate configuration $\config'$. We then sample if this configuration leads to contact by considering the least likely to be $\free$ position $\x_i$ from $ \interiorobs(\config') - \displacement_t$ in Algorithm~\ref{RUMI:alg:dynamics} line~\ref{RUMI:line:sample contact}. We randomly sample from the categorical distribution $s \sim p(S_{\x_i}|\xall)$. If we sample $s = \free$, then the candidate configuration is used as the next one and the object is not displaced. Otherwise, we need to consider if it is a pushing contact. We compute this action's displacement $\displacement'$ by considering the change in position from where $\x_i$ was before the action. Then, we estimate the surface normal $\surfnorm$ at this point in line~\ref{RUMI:line:estimate surface norm} by taking the weighted sum of the SDF gradient of the contact position transformed by each of the pose particles. If the angle between $\surfnorm$ and $-\displacement'$ is less than some threshold $\pushthreshold$ based on an estimation of the friction cone between the robot and the object, then it is considered pushing. If it is a pushing contact, we increase object displacement and move the robot normally. Otherwise, the robot is predicted to remain in its previous configuration to discourage non-pushing contacts, and no further object displacement is produced.

Note that $\config_{t+1}, \displacement_{t+1} \sim \configdynest(\config_t, \displacement_t;...)$ is stochastic since we sample contacts. 
To reduce variance, for a single action trajectory $\us_{\onetohorizon}$ we roll out multiple configuration trajectories by applying $\configdynest$ on copies of the starting configuration $\config_{1}$ and $\us_{\onetohorizon}$. The cost of $\us_{\onetohorizon}$ is the average cost across the multiple $\config_{\onetohorizon+1}$.
Practically, if the object has thin walls relative to the distance a single action could move the robot, as in the case of mugs, each action could be divided up and applied sequentially to avoid dynamics predicting the robot penetrating the object walls.

\begin{algorithm}
\DontPrintSemicolon
\SetKwInput{Given}{Given}
\SetKwInput{Hyperparameters}{Hyperparameters}
\SetKwInOut{Output}{Output}

\Given{
$\config_t$ current robot configuration,\\
$\us_t$ action,\\
$\displacement_t$ current object displacement,\\
$\T_{1..\numparticles}$ pose particles, $\weight_{1..\numparticles}$ particle weights\\
$p(\Sx|\xall)$ semantics probability voxel grid,\\
$\configdyngiven$ given free space dynamics,\\
$\nabla \sdf$ object frame SDF gradient,\\
$\interiorobs$ robot interior points model,\\
$\pushthreshold$ pushing angle threshold
}
\Output{
$\config_{t+1}$ new robot configuration,\\
$\displacement_{t+1}$ new object displacement
}
$\config' \leftarrow \configdyngiven(\config_t, \us_t)$ \tcp{candidate config}
$\interiorset \leftarrow \interiorobs(\config') - \displacement_t$\;
\tcp{find most likely contact}
$i \leftarrow \argmin_{\x_i \in \interiorset} p(S_{\x_i} = \free|\xall)$ \label{RUMI:line:sample contact}\;
$s \sim p(S_{\x_i}|\xall)$ \tcp{sample semantics}
\If{$s = \free$}{
    $\config_{t+1} \leftarrow \config'$\;
    $\displacement_{t+1} \leftarrow \displacement_t$\;
}
\Else{
    \tcp{determine displacement}
    $\interiorset_b \leftarrow \interiorobs(\config_t)$\;
    $\displacement' \leftarrow \x_i - \interiorset_b[i]$\tcp{corresponding ith position}
    \tcp{estimate object surface normal}
    $\surfnorm \leftarrow \sum_{j=1}^\numparticles \weight_j \nabla \sdf(\T_j \x_i)$\label{RUMI:line:estimate surface norm}\;
    \If{angle between $\surfnorm$ and $-\displacement' < \pushthreshold$} { \tcp{pushing or not}
        $\config_{t+1} \leftarrow \config'$\;
        $\displacement_{t+1} \leftarrow \displacement_t + \displacement'$\;
    }
    \Else {
        \tcp{discourage non-pushing contact}
        $\config_{t+1} \leftarrow \config_t$ \; 
        $\displacement_{t+1} \leftarrow \displacement_t$\;
    }
}
\caption{$\configdynest{}$ --- approximate robot and object displacement dynamics}
\label{RUMI:alg:dynamics}
\end{algorithm}

\subsection{Reachability Cost}
For manipulator arms with immobile bases, it is important to explicitly penalize when actions could move the object outside of its reachable region. Under just the information gain cost from Eq.~\ref{RUMI:eq:costinfo}, an action trajectory pushing the object out of reach will evaluate to have equal or better cost than a trajectory doing nothing. 
If the object is at the edge of the robot's reachability, such as a mug with sides that are within reach but the occluded handle at the back being out of reach, sampling a $\nhorizon$ step trajectory that first displaces the mug then collects the high information gain at the back of the mug is very unlikely. $\nhorizon$ may also be too short to allow such a trajectory to exist.

To address this, we introduce \textit{reachability} $\reachability(\x) \in [0, 1]$ and the reachability cost $\costreach(\displacement_{1..\nhorizon})$ which encodes the desired behaviour of pushing object frame points $\x$ with high $\infofield(\x|\xall)$ to where they are reachable.

\begin{figure}
    \centering
    \includegraphics[width=0.49\linewidth]{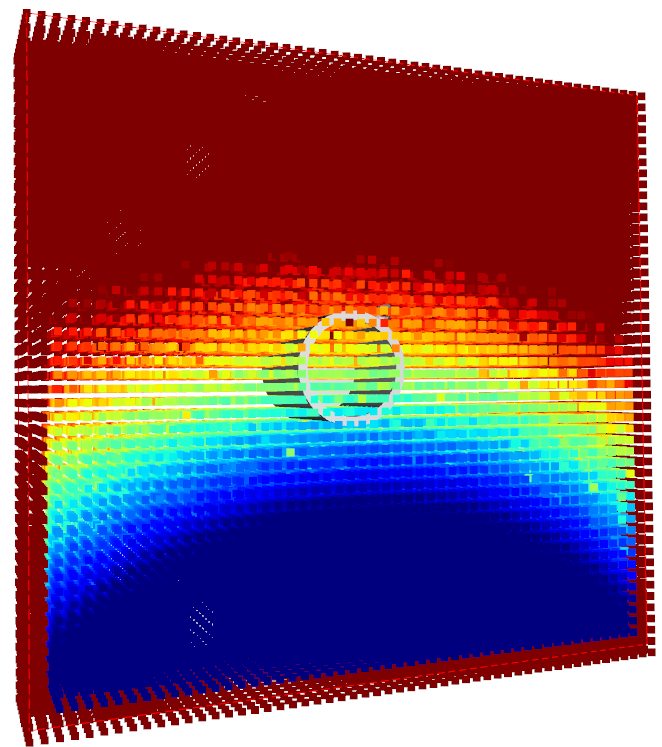}
    \includegraphics[width=0.49\linewidth]{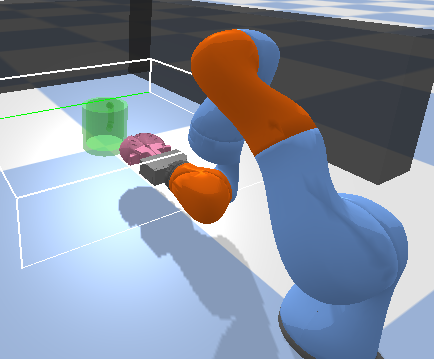}
    \caption{(left) Reachability of workspace positions for the (right) simulated KUKA arm and workspace. Red corresponds to $\reachability(\x)=0$ and dark blue corresponds to $\reachability(\x)=1$. The workspace in sim is drawn as a box around the object.}
    \label{RUMI:fig:reachability}
\end{figure}
\newcommand{\ikrotset}{R}
Reachability $\reachability(\x)$ represents the capability of the robot to gather information at $\x$, similar to checking $\exists \config \ \st \ \x \in \interiorobsinfo(\config)$. This can be approximated by performing inverse kinematics (IK) with $\x$ set as the goal position relative to the robot end effector frame. 
We also consider how robust $\x$ is to reach with different configurations, and evaluate the average IK performance with a fixed set of goal orientations $R_i \in \ikrotset \subset \rotset$. 
Let $\ikerrpos(\x,R_i)$ and $\ikerrrot(\x,R_i)$ be the position and rotation errors from running IK with the goal set to $(\x,R_i)$. We weigh $\ikerrrot$ against $\ikerrpos$ with $\ikscale \geq 0$ and define an error tolerance threshold $\ikthreshold$ such that any error at or above this value receives $\reachability(\x) = 0$. Thus we define
\begin{equation}
    \reachability(\x) = \frac{1}{\ikthreshold}\max(0, \ikthreshold - \frac{1}{|\ikrotset|}\sum_{R_i \in \ikrotset}[\ikerrpos(\x,R_i) + \ikscale\ikerrrot(\x,R_i)])
\end{equation}

We precompute this for $\x \in \workspace$ and store the results in a voxel grid that allows linear interpolation. This only has to be done once per robot and workspace combination. See Fig.~\ref{RUMI:fig:reachability} for an example visualization of $\reachability(\workspace)$.

The reachability cost $\costreach(\displacement_{1..\nhorizon})$ is then the total reachable information within the workspace after displacement. We compute it according to Algorithm~\ref{RUMI:alg:reachable cost}. First we compute $\bar{\infofield}(\x)$, the average information gain at every displaced workspace position over the planning horizon. Note that $\infofield(\x|\xall)$ can be interpreted as the object frame \infofieldname{} at the time of planning, and so stationary workspace positions are effectively displaced by $-\displacement_{1..\nhorizon}$ during planning. The reachable information is just the product $\bar{\infofield}(\x)\reachability(\x)$ which we sum across all the workspace points. This is then compared against the total information in the workspace to produce a negative ratio $\costreach \in [-1, 0]$. Because $\costinfo$ and $\costreach$ are in different units, having $\costreach$ be a ratio allows easier tuning of the total trajectory cost:
\begin{equation}\label{RUMI:eq:mppi cost}
    \totalcost(\config_\onetohorizon, \displacement_\onetohorizon) = \scaleinfo\costinfo(\config_\onetohorizon, \displacement_\onetohorizon) + \scalereach\costreach(\displacement_\onetohorizon)
\end{equation}

\begin{algorithm}
\DontPrintSemicolon
\SetKwInput{Given}{Given}
\SetKwInput{Hyperparameters}{Hyperparameters}
\SetKwInOut{Output}{Output}

\Given{
$\displacement_{1..\nhorizon}$ object displacement trajectory,\\
$\workspace$ workspace,\\
$\infofield(\x|\xall)$ information gain voxel grid,
}
$\bar{\infofield}(\x) \leftarrow \frac{1}{\nhorizon} \sum_{t}^{\nhorizon} \infofield(\workspace - \displacement_t | \xall)$ \tcp{average info}
$RI \leftarrow \sum_{\x \in \workspace} \bar{\infofield}(\x) \reachability(\x)$ \tcp{reachable info}
$RI_m \leftarrow \sum_{\x \in \workspace} \infofield(\x|\xall)$ \tcp{max info possible}
$\costreach(\displacement_{1..\nhorizon}) \leftarrow - RI / RI_m$\;
\caption{$\costreach{}$ reachability cost}
\label{RUMI:alg:reachable cost}
\end{algorithm}

\subsection{Kernel Interpolated MPPI}
The total cost from Eq.~\ref{RUMI:eq:mppi cost} does not include any explicit smoothing terms. To improve the smoothness of produced trajectories, we perform interpolation similar to~\cite{miura2024spline}. The idea is to sample $\nhorizoninterp < \nhorizon$ control points $\controlpts_i \in \reals^\nus$, then use a kernel $\kernel: \reals^\nus \times \reals^\nus \rightarrow \reals$ to interpolate the $\us_\onetohorizon$ in between $\controlpts_\onetointerp$. We call this method Kernel Interpolated MPPI (KMPPI). This is more general than the B-spline interpolation of~\cite{miura2024spline} since it can be accomplished by using a B-spline kernel.

\begin{figure}
    \centering
    \includegraphics[width=\linewidth]{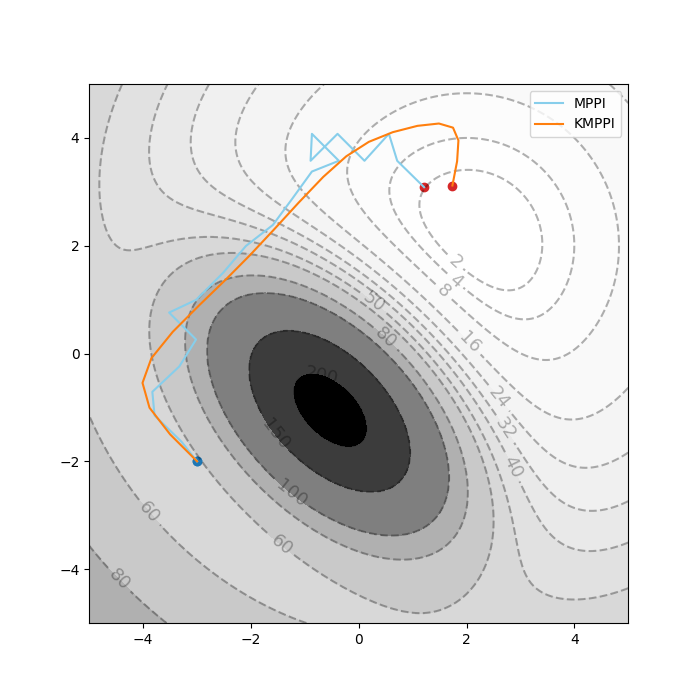}
    \caption{Planned trajectories on a toy 2D linear integrator environment. The cost contour map is represented, with the cost of the trajectory being the accumulated cost experienced at each state. There is additionally a quadratic action penalty at each step $\us^T 0.1 \mathbf{I} \us$. Dark blue is the starting state. Each trajectory has $\nhorizon=20$, with our KMPPI having $\nhorizoninterp=5$ and using the radial basis function kernel with scale 2.}
    \label{RUMI:fig:kmppi}
\end{figure}

Let $\tuorig = [0, 1, ..., \nhorizon - 1]$ denote the time coordinate of each $\us$ along the trajectory. We assume $\controlpts_\onetointerp$ are evenly spread out along the trajectory, and since there are $\nhorizoninterp$ of them, subsequent ones increase their time coordinate by $\frac{(\nhorizon-1)}{(\nhorizoninterp-1)}$ to give $\tuinterp = [0, \frac{(\nhorizon-1)}{(\nhorizoninterp-1)}, \frac{2(\nhorizon-1)}{(\nhorizoninterp-1)}, ..., \nhorizon-1]$. The even assignment of $\tuinterp$ is not necessary; any can be given as long as the first term is 0 and the last term is $\nhorizon-1$. Given a control sequence $\controlpts_\onetointerp$ we then convert it to $\us_\onetohorizon$
\begin{align}
    \us_\onetohorizon &= \kernel(\tuorig, \tuinterp)\kernel(\tuinterp, \tuinterp)^{-1} \controlpts_\onetointerp
\end{align}
This allows smoothing in the action space, rather than in the robot configuration space, and we observe that it works well on our tasks. See Fig.~\ref{RUMI:fig:kmppi} for a qualitative evaluation of the smoothing property on a toy 2D problem.

With the interpolated $\us_\onetohorizon$, KMPPI's subsequent steps are the same as MPPI's in that it generates configuration rollouts by applying the dynamics function $\config_{t+1}, \displacement_{t+1} \sim \configdynest(\config_t, \displacement_t;...), t=\onetohorizon$, evaluates the cost of each $\us_\onetohorizon$ with Eq.~\ref{RUMI:eq:mppi cost}, then combines the trajectory samples with a softmax based on the cost.

\subsection{Termination Condition}
In actual execution, we do not have access to $\Top$ to evaluate $\nll(\xall)$ and need another signal to terminate execution. We use the convergence of the pose particles, with the hypothesis that pose particles likely only converge when $p(\Top|\xall)$ is high, i.e. the pose particles do not randomly converge to an incorrect estimate. We evaluate convergence using the average square root pairwise Chamfer distance between the pose particles ($\avgpairwisechamfer$). Similar to the $\nll(\xall)$ evaluation, we evaluate this on a sampled set of object frame surface positions $\objframept \in \objframeset$.
\begin{equation}\label{RUMI:eq:avgchamfer}
    \avgpairwisechamfer(\T_\onetoparticles) = \frac{1}{\numparticles^2 |\objframeset|} \sum_{i=1}^\numparticles \sum_{j=1}^\numparticles \sum_{\objframept \in \objframeset} |\sdf(\T_i^{-1}\T_j\objframept)|
\end{equation}
We terminate execution when $\avgpairwisechamfer(\T_\onetoparticles) < \chamfertermscale \characteristiclength$, where $\characteristiclength$ is the diagonal length of the object's  bounding box, and $\chamfertermscale$ is a ratio that selects for a desired level of pose particle convergence. A lower value means rummaging will continue for longer, but may produce a more accurate pose estimate.
\section{Experiments}
\label{RUMI:sec:results}
In this section, we first describe our simulated and real robot environments. We then detail the experiments to estimate the pose of a movable target object. We introduce our baselines and ablations and how we quantitatively evaluate the methods on the experiment. \rev{We then show how \rumiabv{} could be applied when there are an unknown number of unknown movable objects cluttered around our target object.} Lastly, we present results that show \rumiabv{} is the only method to perform consistently well across all the experiments.

\subsection{Sim Environment}
Common to all the experiments, we have a single movable object on a flat surface starting within reach of a single 7DoF KUKA LBR iiwa arm with two soft-bubble tactile sensors~\cite{kuppuswamy2020soft}. This is modelled in sim in Fig.~\ref{RUMI:fig:reachability}. Due to the complexity of modelling deformable objects, we model the soft-bubble tactile sensors as rigid bodies and observe the surface points of any object penetrating them after each simulation step. We also include a fixed external depth camera to reduce the initial exploration required, but we also show that our method works without a good initial view of the object in some experiments. 

\begin{figure}
    \centering
    \includegraphics[width=0.6\linewidth]{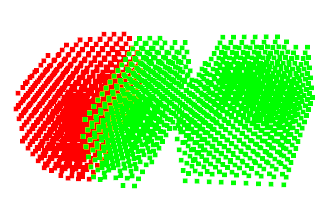}
    \caption{Visualization of interior robot points $\interiorobs(\config)$ as green and red points, and the information gathering subset $\interiorobsinfo(\config)$ as just the red points for the robot gripper mounted with two soft-bubble tactile sensors seen in Fig.~\ref{RUMI:fig:teaser}.}
    \label{RUMI:fig:info gathering mask}
\end{figure}

For observing $\s=\surface$ points at $\config_t$, we select $\{\x | \ \x \in \interiorobsinfo(\config_t), |\sdf(\Top \x)| \ < 3mm\}$. 
This simulates some observation noise which we show that \rumiabv{} is robust to, despite assuming no noise in the observed positions. We assume we can only gather contact information from the front of the gripper, where the two soft-bubble tactile sensors are mounted. In planning, this is the difference between $\interiorobs(\config)$ and $\interiorobsinfo(\config)$ for our end effector shown in Fig.~\ref{RUMI:fig:info gathering mask}.

\begin{figure}
    \centering
    \includegraphics[width=0.49\linewidth]{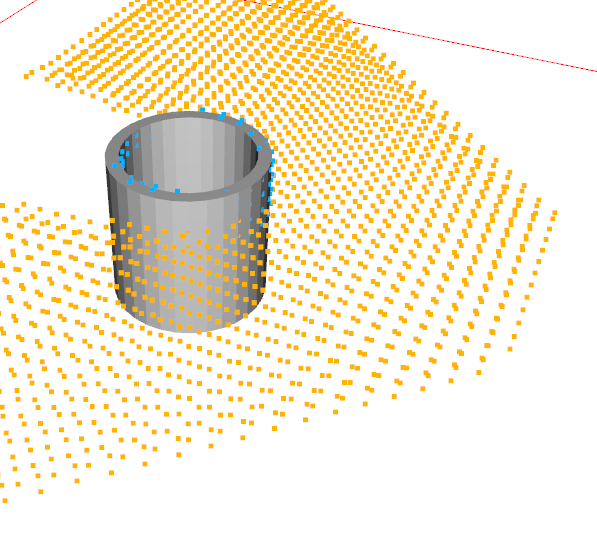}
    \includegraphics[width=0.49\linewidth]{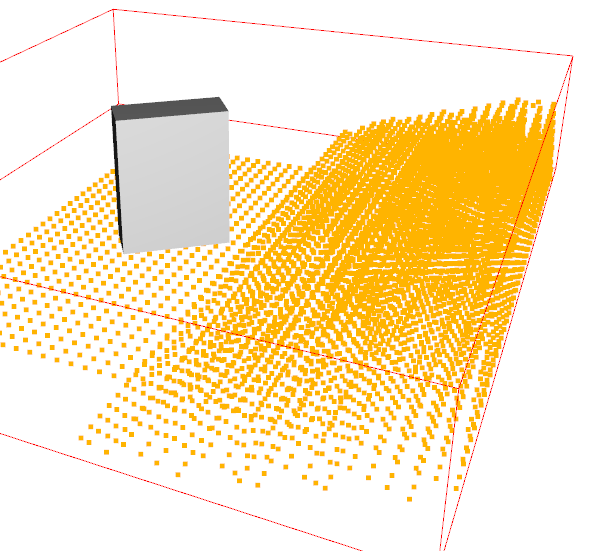}
    \caption{Comparison of the rendered $\xall_0$ given to (left) the sim mug 0 task and (right) sim box 2 task. $\s=\free$ points are in orange, and $\s=\surface$ points are in blue. There are no initially observed surface points on the box.}
    \label{RUMI:fig:init render}
\end{figure}

The $\xthis$ provided by the depth camera includes $\s = \free$ points generated by tracing rays from the camera to 95\% of each pixel's detected depth, and $\s = \surface$ from segmented object surfaces. See Fig.~\ref{RUMI:fig:intuition} (middle), and Fig.~\ref{RUMI:fig:init render} for example observation point clouds. We only use vision to provide the initial $\xall_0$ to demonstrate the viability of tactile based rummaging. 

To highlight the difference between other components of all methods, we directly observe $\Delta \T$ in Algorithm~\ref{RUMI:alg:update} line~\ref{RUMI:line:objmove}. Note that this also applies to all baselines and ablations, and so does not provide an unfair advantage to \rumiabv{}. This is equivalent to assuming we can accurately measure slip between the end effector and object. 

\begin{figure}
    \centering
    \includegraphics[width=\linewidth]{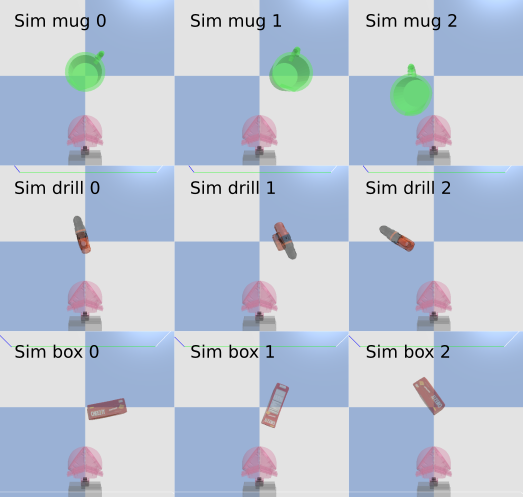}
    \caption{Starting poses for labelled sim tasks.}
    \label{RUMI:fig:sim rmg init}
\end{figure}

\subsection{Sim Tasks}
In simulation, we experiment on 3 different objects: a mug, a YCB~\cite{calli2017yale} power drill, and a YCB cracker box, each with 3 different initial poses depicted in Fig.~\ref{RUMI:fig:sim rmg init}. 
For each, we perform 10 runs of $\nsteps=40$ steps, using a different fixed random seed for each run that is shared across baselines and ablations. We terminated tasks early if the pose particles converged as measured by $\avgpairwisechamfer(\T_\onetoparticles) < 0.03 \characteristiclength$, where $\characteristiclength$ is the diagonal distance of each object's bounding box.

In each experiment, the robot's end effector is position and yaw controlled, with the action space either being $\us = [dx, dy, d\theta]$ (planar) or $\us = [dx, dy, dz, d\theta]$ (3D) with ranges from [-1,1] for each dimension. The action spaces are scaled to allow the use of consistent KMPPI parameters across experiments. We scale these to physical units by translating a control value of 1 to $80mm$ or $0.5$ radians, carried out in many mini steps. We perform inverse kinematics to convert these to joint position commands. We used regular grids with resolutions (grid square side length) $\resworkspace$ as the workspaces. Note that other sets of worldspace points that are not necessarily regular grids could be used. We used $\workspace = [0, 0.8] \times [-0.4, 0.4]$ in meters for planar action spaces, and  $\workspace = [0, 0.8] \times [-0.4, 0.4] \times [0, 0.2]$ for 3D action spaces. $\resworkspace$ for each task can be found in Tab.~\ref{RUMI:tab:objects}. 

Each object is intended to illustrate a different aspect of exploration. For the mug and power drill, we assume the object stays upright and search for their pose in $\posesetplanar$ instead of $\poseset$. The mug tasks evaluates how well $\infofield(\x|\xall)$ conforms to our intuition, since we expect the most information to be where the handle could be. The sim drill task evaluates how well our planner extends to objects with complex geometry. The sim box task tests how well the pose particles can represent full $\poseset$ and the necessary 3D exploration to identify which side of the box is lying against the floor. Additionally, for the drill and box tasks, we increase the difficulty in terms of environmental occlusions by placing the camera at an angle such that it cannot directly observe the object. The camera configurations and the initial object pose are depicted in Fig.~\ref{RUMI:fig:init render}. For $\poseset$ pose search in the sim box experiments, we add free points where the floor is to avoid pose estimates that penetrate the floor. The different task setups are summarized in Tab.~\ref{RUMI:tab:objects}.

We use different $\Tprior$, the prior pose particles, for the mug tasks where we initially observe the front of it, to the other tasks where we initially cannot see it. For mugs, we first estimate the position of the center of the mug, then $\Tprior$ is sampled with uniformly random yaw and the same center. For the other tasks, we sample $\Tprior$ with random positions sampled from $\mathcal{N}(0,0.05) \times \mathcal{N}(0,0.05) \times 0$, and also uniformly random yaw (assuming upright).

\subsection{Real Environment}
The real robot setup is seen in Fig.~\ref{RUMI:fig:teaser} and Fig.~\ref{RUMI:fig:sim rmg init}. It uses the same robot as in simulation. The soft-bubble sensors are compliant to contact and have a depth camera inside to estimate dense contact patches. Similar to prior work~\cite{zhong2023chsel}, we consider points on the soft bubble surface with deformation beyond $4mm$ and being in the top $10^{th}$ percentile of all deformations to be in contact. We apply a mean filter to remove noise. We use a RealSense L515 lidar camera as the fixed external camera. For evaluating ground truth object pose, we have a RealSense D435 camera mounted looking top-down on the workspace.

The mug had distinct colors from the shelf and so we segmented it with a color filter. To improve segmentation, we used a robot self-filter and an edge filter to remove unreliable points, and used a temporal filter to only accept surface points that persists over a 0.4s window. See Fig.~\ref{RUMI:fig:teaser} for example observation point clouds. We re-observe the scene after each action. Due to self-occlusion and object symmetry, visual observations do not uniquely identify object pose. Same as for the simulated box, the real box task has occluded vision that prevented direct observation of it, seen in the top of Fig.~\ref{RUMI:fig:real rmg init}.

For the real mug task, we do not assume we can accurately measure slip between the end effector and object. Instead, we estimate $\Delta \T$ with Algorithm~\ref{RUMI:alg:objmove} for all methods. For the real box task, we observe the change in object pose from the ground truth since we cannot directly observe the object to estimate $\Delta \T$ with Algorithm~\ref{RUMI:alg:objmove}.

\subsection{Real Task}
\begin{figure}
    \centering
    \includegraphics[width=\linewidth]{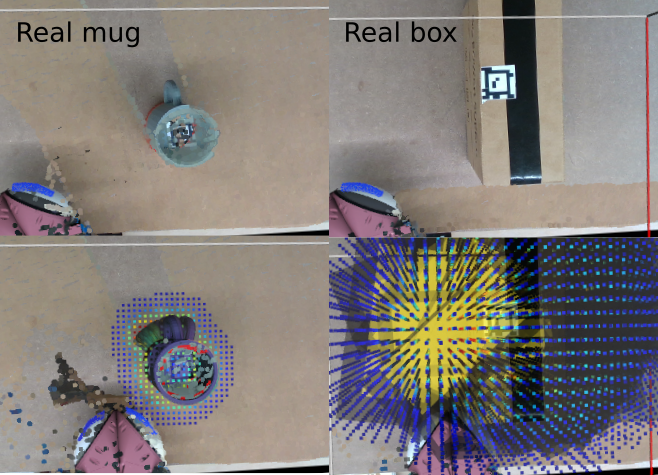}
    \caption[Initial configuration of the real mug and box tasks and example initialized pose particles]{Initial configuration of the real mug and box tasks (top) and example initialized pose particles (bottom). In the box task, the workspace is occluded and the box cannot be directly observed. Point cloud observations from an external side view (accessible to the robot) are overlaid on a top-down raw camera view (inaccessible to the robot).}
    \label{RUMI:fig:real rmg init}
\end{figure}

We estimate the pose of a real mug and box starting in a single configuration depicted in Fig.~\ref{RUMI:fig:real rmg init} and execute $\nsteps=15$ steps of each method. The robot's action space is seen in Tab.~\ref{RUMI:tab:objects}, and a control value of 1 corresponds to $50mm$ or $0.4$ radians. The workspace was $\workspace = [0.55, 1.1] \times [-0.33, 0.33] \times [0.23, 0.47]$ in meters (for planar action space, a fixed height of 0.305m was used). The $\Tprior$ initialization process is similar to sim for each corresponding task, with $\T_\onetoparticles$ after sampling from CHSEL in Algorithm~\ref{RUMI:alg:update} line~\ref{RUMI:line:init} shown at the bottom of Fig.~\ref{RUMI:fig:real rmg init}. Note that the box's initial $\T_\onetoparticles$ covers the workspace since vision was occluded.
We terminated tasks when the pose particles converged as measured by $\avgpairwisechamfer(\T_\onetoparticles) < 0.05 \characteristiclength$, where $\characteristiclength$ is the diagonal distance of each object's bounding box.

\begin{table}
\centering
\begin{tabular}{|l|l|l|l|}
\hline
Object     & action space & pose search space & resolution $\resworkspace$ (m) \\ \hline
Sim mug    & planar       & SE(2) & 0.01            \\
Sim drill  & 3D           & SE(2) & 0.02            \\
Sim box    & 3D           & SE(3) & 0.02            \\
Real mug   & planar       & SE(2) & 0.01            \\
Real box   & 3D           & SE(2) & 0.02            \\ \hline
\end{tabular}
\caption{Task setup for different objects.}\label{RUMI:tab:objects}
\end{table}

\newcommand{\sdfval}{v}
\newcommand{\modifiedsdfval}{\Tilde{v}}
\newcommand{\sensorexp}{\alpha}
\newcommand{\sensorthreshold}{\zeta}
\newcommand{\discountfactor}{\beta}
\subsection{Sensor Model}
We use the sensor model depicted in Fig.~\ref{RUMI:fig:sensor}. Let $\sdfval=\sdf(\T\x)$ be the SDF value of a given query position $\x$. 
To represent bias towards over-reporting contact in our sensors, we use a tolerance of $\sensorthreshold=0.003m$ and let $\modifiedsdfval = \sign(\sdfval) \cdot \max(0, |\sdfval| - \sensorthreshold)$, where $\sign(\sdfval)=1$ if $\sdfval > 0$ else $-1$. 
Then $p(\Sx|\T) = p(\Sx|\sdf(\T\x)) = p(\Sx|\sdfval)$ is defined by
\begin{align*}
p(\Sx|\sdfval) = \begin{cases} \max(0, 1 - e^{-\sensorexp \modifiedsdfval}) & \free \\
\max(0, 1 - e^{\sensorexp \modifiedsdfval})  & \occupied \\
e^{-\sensorexp |\modifiedsdfval| } & \surface \end{cases}
\end{align*}
with $\sensorexp = 100$ where $\sdfval$ is in meters. This model represents some of the ambiguities of detecting contact with the soft-bubble and similar tactile sensors. Due to the compliance of the membrane, even when a point is in free space, contact elsewhere could make it appear like this point is also in contact. Similarly, contact could also be missed, particularly around the edges of the soft-bubble. This sensor model performed well enough both in sim and on the real task that no calibration to the real soft-bubbles was needed.

\subsection{KMPPI Parameters}
We used a planning horizon of $\nhorizon = 15$ and $\nhorizoninterp = 8$ number of control points. This is lower than the number of sim steps $\nsteps = 40$ because increasing horizon resulted in poorer-quality trajectories. This is due to the cost from Eq.~\ref{RUMI:eq:mppi cost} being a terminal cost for the whole trajectory, without distinguishing between steps inside the trajectory. We used the radial basis function (RBF) kernel with a scale of 2.

We planned using 500 action trajectory samples, each rolled out 5 times with $\configdynest$ due to its stochastic nature. Additionally, to avoid contacts that penetrate the object, we split each action up into 4 sequentially applied actions that are 4 times lower in magnitude. We then use the average trajectory cost across the 5 rollouts. We replanned after executing 3 actions, or when the robot detects it is in contact.

For the inner MPPI parameters, we used $\lambda = 0.01$ for the temperature parameter from~\cite{williams2017model}, with $\mathbf{0}$ noise mean and $1.5\mathbf{I}$ as the noise covariance.

\subsection{Evaluation}
\rev{
We evaluate using the key metric of NLL defined in Eq.~\ref{RUMI:eq:nll}. Note that low NLL indicates both certainty and correctness of the pose distribution.
}
We sample 500 positions $\objframept \in \objframeset$ uniformly on the surface of the object, and transform them to world positions with $\Top$, the ground truth pose, to produce $\X$.
We then evaluate the negative log likelihood of $\x \in \X$ being $\surface$ points from Eq.~\ref{RUMI:eq:nll}. Because we assume $\Sx$ is conditionally mutually independent to every other $\Sx$ given $\xall$, we can simplify Eq.~\ref{RUMI:eq:nll}
\begin{align}
      \nll(\xall) &= - \log p(\bigcap_{\x \in \X} \Sx = \surface | \xall)\\
      &=  - \sum_{\x \in \X} \log p(\Sx = \surface | \xall)
\end{align}
We substitute Eq.~\ref{RUMI:eq:calcsem} in for $p(\Sx|\xall)$ to approximate $\nll$ with our pose particles
\begin{align}
    \nll(\xall) &\approx - \sum_{\x \in \X} \log \sum_{i=1}^{\numparticles}  \weight_i p(\Sx = \surface | \sdf(\T_i \x))
\end{align}

For the sim tasks, we have the ground truth object pose $\Top$, while for the real tasks, we observe $\Top$ from a camera mounted above the workspace. We evaluated $\nll$ after each step as an effective exploration rate. Additionally, we specify a $\nll$ threshold below which we qualitatively observe to be a good enough quality to be considered a success, seen in Tab.~\ref{RUMI:tab:nll}. A run is counted a success if it achieves a minimum $\nll$ below the threshold at any step. This is typically, but not always, the last step. This is because, due to observation noise and moving the object outside of the observed region, the pose estimates could become less certain.

We also use the same $\objframept \in \objframeset$ to evaluate $\avgpairwisechamfer$ from Eq.~\ref{RUMI:eq:avgchamfer}. We used $\chamfertermscale = 0.03$ for the sim tasks, meaning we terminated exploration when the average square root chamfer distance between all pairs of $\T_\onetoparticles$ is less than 3\% of the object's bounding box diagonal length. For the real experiment we used $\chamfertermscale = 0.05$.

\begin{figure}
    \centering
    \includegraphics[width=\linewidth]{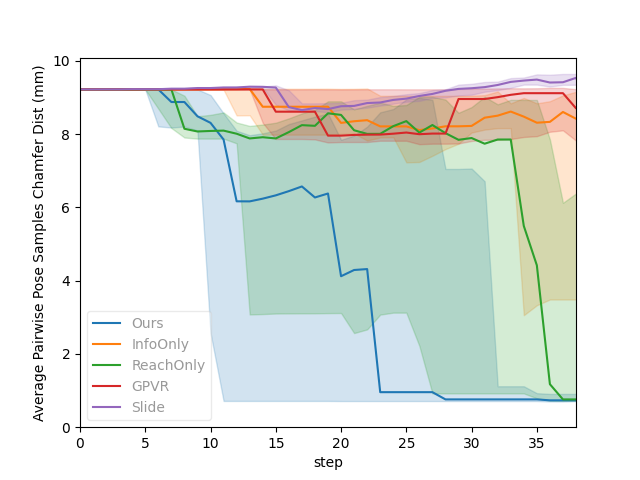}
    \includegraphics[width=\linewidth]{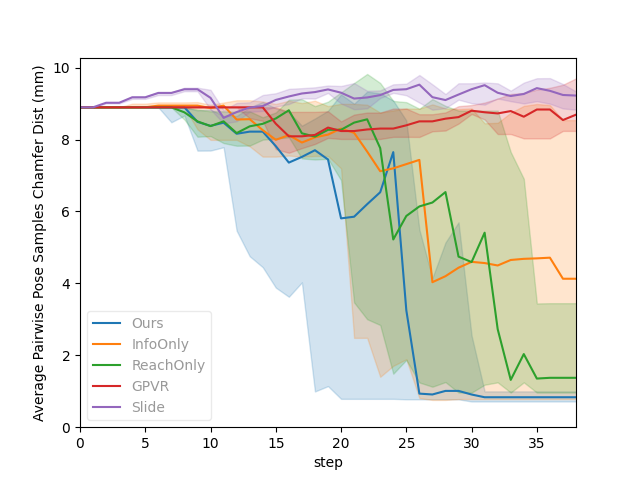}
    \caption{Convergence of pose particles measured by $\avgpairwisechamfer(\T_\onetoparticles)$ for the sim mug 0 and sim mug 1 tasks. The median over 10 runs is plotted, with the $25^{th}$ to $75^{th}$ percentile shaded. There is strong correlation with the $\nll(\xall)$ in the top left and top middle of Fig.~\ref{RUMI:fig:rmg sim res}.}
    \label{RUMI:fig:convergence}
\end{figure}

We also investigated our hypothesis of $\avgpairwisechamfer(\T_\onetoparticles)$ as a good proxy for $\nll(\xall)$ since it can be computed without privileged information. We did so by computing the linear correlation between the two across all the tasks. The runs from all methods were used. This is shown in Tab.~\ref{RUMI:tab:convergence} and Fig.~\ref{RUMI:fig:convergence} for the sim mug 0 and sim mug 1 tasks, which can be compared to the $\nll(\xall)$ shown in the top left and top middle of Fig.~\ref{RUMI:fig:rmg sim res}. We see that there is an especially strong positive correlation for $\posesetplanar$ particles of the sim mug and sim drill tasks, averaging to a correlation of \textbf{0.87}. The correlation for the $\poseset$ sim box tasks is not as strong.

\begin{table}
\centering \setlength{\tabcolsep}{3pt}
\begin{tabular}{|l|lll|lll|lll|}
\hline
\multirow{2}{*}{Task} &
  \multicolumn{3}{l|}{sim mug} &
  \multicolumn{3}{l|}{sim drill} &
  \multicolumn{3}{l|}{sim box} \\ \cline{2-10} 
 &
  0 &
  1 &
  2 &
  0 &
  1 &
  2 &
  0 &
  1 &
  2 \\ \hline
cor($\nll,\avgpairwisechamfer$) &
  \multicolumn{1}{r}{0.93} &
  \multicolumn{1}{r}{0.82} &
  \multicolumn{1}{r|}{0.84} &
  \multicolumn{1}{r}{0.74} &
  \multicolumn{1}{r}{0.94} &
  \multicolumn{1}{r|}{0.94} &
  \multicolumn{1}{r}{0.58} &
  \multicolumn{1}{r}{0.80} &
  \multicolumn{1}{r|}{0.49} \\ \hline
\end{tabular}
\caption{Linear correlation between $\nll(\xall)$ and $\avgpairwisechamfer(\T_\onetoparticles)$ across all sim tasks. Runs from all methods were considered in its calculation.}\label{RUMI:tab:convergence}
\end{table}

\begin{table}
\centering
\begin{tabular}{|l|l|}
\hline
Object    & success $\nll$ threshold \\ \hline
Sim mug    & 20       \\
Sim drill  & 100           \\
Sim box    & 150           \\
Real mug   & 35       \\ 
Real box   & 250       \\ 
\hline
\end{tabular}
\caption{Maximum $\nll$ threshold for success for each task.}\label{RUMI:tab:nll}
\end{table}

\subsection{Baselines and Ablations}
Our full method parameters are summarized in Tab.~\ref{RUMI:tab:params}. These parameters were used for all simulated and real tasks (except for $\chamfertermscale$ in deciding when to terminate), demonstrating the robustness of \rumiabv{}. For downsampling the observations in Algorithm.~\ref{RUMI:alg:mergeobs}, we used different resolutions for the free space ($\resdownsamplefree$) and surface ($\resdownsamplesurface$) points; we did not observe any occupied points. The baselines also required the creation and update of the $p(\T|\X)$ pose particles, and we use the same parameters to do so.

We present two ablations to our full method, \textbf{\infoonly{}} which sets $\costreach$ to 0 and \textbf{\reachonly{}} which sets $\costinfo$ to 0. They share all other parameters with the full method and evaluate the usefulness of each individual cost.

For baselines, we first present the \textbf{\slide{}} heuristic inspired by~\cite{driess2017active}. This method has two modes of operation - if it is currently in contact, then it moves tangentially to the estimated surface normal to slide along it. It moves parallel to the shelf, and for each run randomly decides at the start of the run whether to slide clockwise or counterclockwise around contact. If it is not in contact, then it moves towards the estimated center of the object. Estimating the object center requires our pose particles, so we still update $p(\T|\xall)$ using Algorithm~\ref{RUMI:alg:update}. 

\begin{figure}
    \centering
    \includegraphics[width=0.49\linewidth]{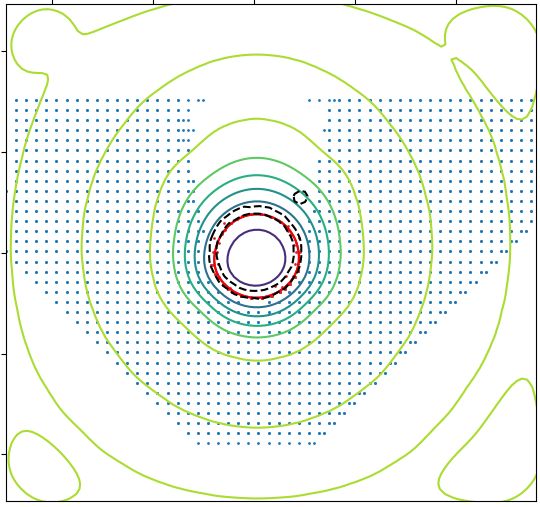}
    \includegraphics[width=0.49\linewidth]{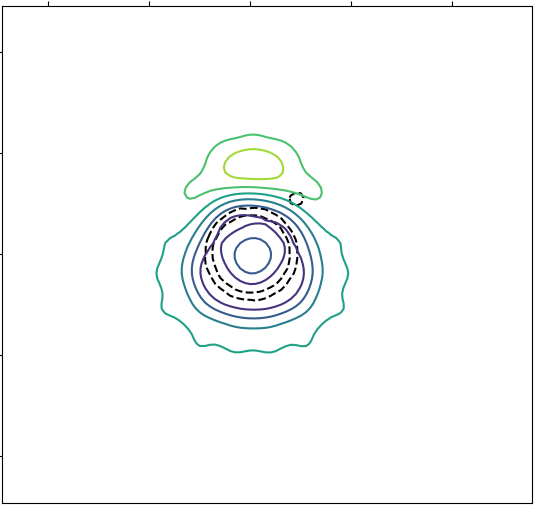}
    \caption{(left) Gaussian Process fit to initial observations $\xall_0$ of the sim mug 0 task. Free space observed points are shown as blue dots, surface observed points as red dots, and the ground truth object surface as black dotted lines. The GP output is overlayed as a contour map, with the red line indicating the 0-level set, corresponding to where the GPIS surface is. (right) $\var(\x|\X)$ contour map for the same GP, with green indicating higher variance.}
    \label{RUMI:fig:gp example}
\end{figure}

We also consider a Gaussian Process Implicit Surface baseline (GPIS)~\cite{caccamo2016active},~\cite{driess2017active},~\cite{lee2019online} that uses the variance of the GP as the exploration signal that we call GP Variance Reduction (\textbf{\gpvr{}}). The GP is fit on $\{(\x, 0) | \ (\x,\s) \in \xall_t, \s = \surface \} \cup \{(\x, 1) | \ (\x,\s) \in \xall_t, \ \s = \free \} $. As typical for GPIS, surface points are labelled 0 and free points are labelled 1. 
It is refit on $\xall_t$ for 50 optimization steps after every robot execution step. We use the $\nu=1.5$ Matern kernel as recommended by~\cite{lee2019online}. See Fig.~\ref{RUMI:fig:gp example} for a visualization of the fitted GP as well as its variance $\var(\x|\xall)$ on the sim mug 0 task given $\xall_0$.

For \gpvr{} to be competitive, we had to make several modifications. 
Firstly, we needed to encode object shape as that is given information to \rumiabv{}. This is non-trivial to do by modifying the kernel, so we instead augmented the input data with $\{(\x,1)| \ \x  \in \workspace, \ p(\Sx = \free|\xall_t) > 0.99\}$. 
We voxel downsampled all free points with a resolution 7 times $\resworkspace$ from Tab.~\ref{RUMI:tab:objects} to avoid extremely slow inference and enforce consistent data density.
Again, this baseline requires the computation and maintenance of $p(\T|\xall)$ with the pose particles to enable the estimation of $p(\Sx = \free|\xall_t)$. Without the above data augmentation, \gpvr{} explores the unobserved corners of the workspace, despite seeing parts of the object elsewhere. Secondly, we needed to plan further than just the next step. Otherwise, because we start in and are surrounded by free space, the method goes in initially random directions. Instead of the greedy policy of maximizing the GP variance $\var(\x)$ at the next position from~\cite{driess2017active}, 
we formulated a cost function based on variance reduction for use as a running cost inside KMPPI.
\begin{equation}\label{RUMI:eq:costgp}
    C_{GP}(\config_t, \displacement_t) = \sum_{\x \in \downsample(\interiorobsinfo(\config_{t}) - \displacement_t,\resdownsample) } -  \discountfactor^{t-1}\var(\x | \xall)
\end{equation}
with a discount factor $\discountfactor=0.99$ to prioritize early rewards. Empirically, this worked better with $\var(\x|\xall)$ than voxelizing the entire trajectory as in Eq.~\ref{RUMI:eq:costinfo}. Before each planning step we precomputed $\var(\x|\xall) \ \forall \x \in \workspace$ to store in a voxel grid for faster repeated lookup. We normalized $\var(\x|\xall)$ such that $\max_{\x \in \workspace} \var(\x|\xall) = 1$.


See Fig.~\ref{RUMI:fig:gpvr vs ours} for a comparison of $\infofield(\x|\xall)$ against $\var(\x|\xall)$ to be planned over in a similar manner.
From the figure, we see that  $\var(\x|\xall)$ is low at where the handle could be. This is because $\xall$ includes the inside back of the mug, and the Matern kernel does not directly encode object shape but is just based on the Euclidean distance between points. It cannot separate the certainty of the back surface of the mug from the uncertainty of where the handle is, because it does not know that a handle exists.
Instead, $\var(\x|\xall)$ is highest farther behind the mug, where we have observed no data due to occlusion.
This is contrasted with $\infofield(\x|\xall)$, which is highest where the handle could be because those regions are where the pose particles disagree the most.

\begin{table}
\centering
\begin{tabular}{|l|l|}
\hline
Parameter     & value \\ \hline
$\numparticles$ number of pose particles & 100 \\
$\peakiness$ peakiness  & 2 \\
$\resamplethreshold$ discrepancy resample threshold  & 5 \\
$\scalefree$ CHSEL freespace discrepancy scale & 10 \\
$\nhorizon$ planning horizon  & 15 \\
$\pushthreshold$ pushing angle threshold  & 45 degrees \\
$\costinfo$ information gain cost scale  &  1 \\
$\costreach$  reachability cost scale &  200 \\
$\ikthreshold$ reachability IK error threshold & 0.4 \\
$\ikscale$ reachability IK rotation error scale & 0.1 \\
$\resdownsamplefree$ downsample resolution free space  &   10mm   \\
$\resdownsamplesurface$ downsample resolution surface  &  2mm  \\
$\updatenoiset$ pose translation noise   &  10mm    \\
$\updatenoiser$ pose rotation noise   &  0 \\
$\chamfertermscale$ chamfer distance convergence ratio   &  0.03 (0.05 for real) \\
$\noptim$ number of optimization steps    & 10 \\ \hline
\end{tabular}
\caption{Our full method parameters across the different tasks.}\label{RUMI:tab:params}
\end{table}

\begin{figure}
    \centering
    \includegraphics[width=\linewidth]{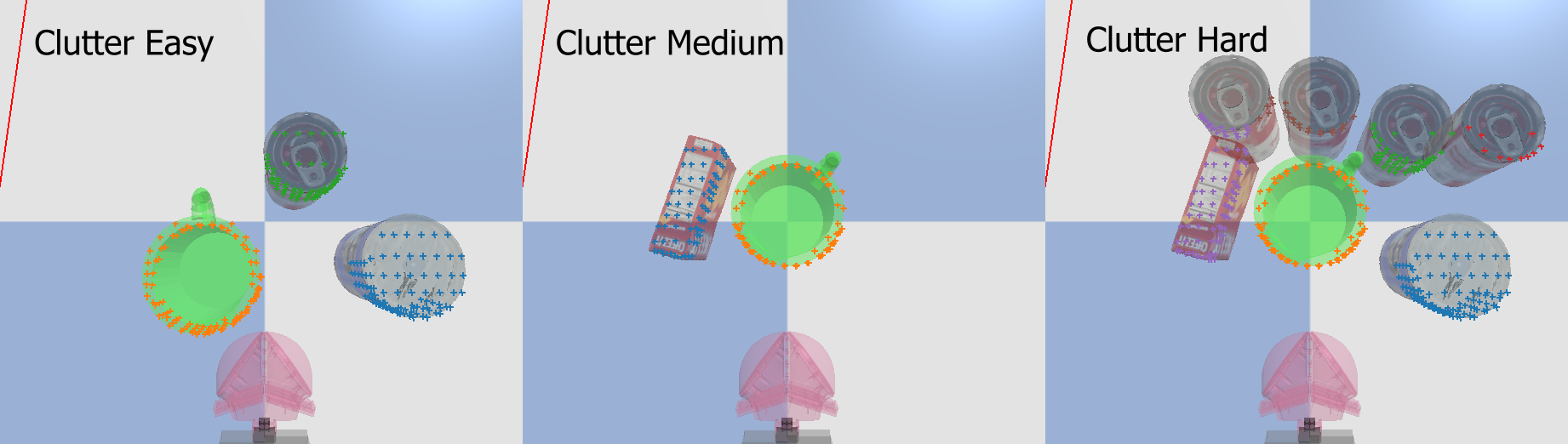}
    \caption{\rev{Starting configuration of clutter experiments with clustered surface points rendered in different colors. All objects are movable and the mug is the target object. Clutter Medium may appear simpler than Clutter Easy, but due to the closer proximity of the objects, it becomes a more difficult task.}}
    \label{RUMI:fig:clutter start}
\end{figure}

\rev{
\subsection{Rummaging in Clutter Experiment}\label{RUMI:sec:clutter}
To evaluate the applicability of \rumiabv{} in more practical settings, we consider sim tasks that have an unknown number of unknown movable objects cluttered around the target object depicted in Fig.~\ref{RUMI:fig:clutter start}. 
}

\begin{figure}
    \centering
    \includegraphics[width=\linewidth]{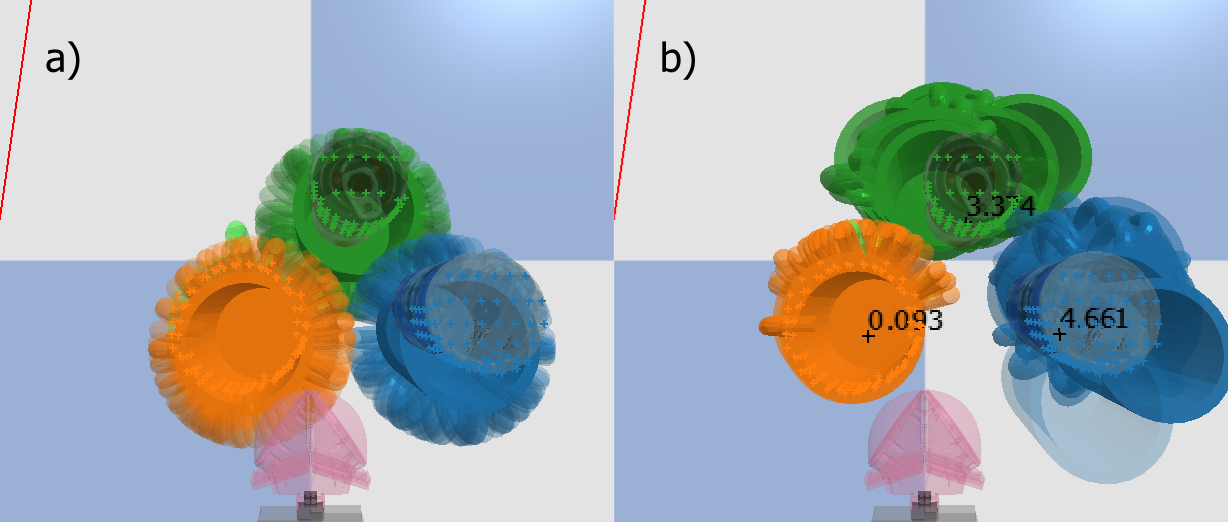}
    \caption{\rev{Target \objcluster{} selection process on Clutter Easy with a) 100 initial transforms centered on the centroid of each $\xcluster$ and b) those transforms after applying CHSEL and the resulting median discrepancy of each cluster.}}
    \label{RUMI:fig:select target}
\end{figure}

\rev{
To handle the presence of other objects, we first cluster the initial camera observation's surface points using HDBSCAN~\cite{mcinnes2017hdbscan} into \nclusters{} \textit{\objcluster{}s}. \nclusters{} is not specified but instead determined by HDBSCAN. Let $\xcluster,  k=1..\nclusters{}$ denote each \objcluster{}, with every point having \surface{} semantics. As short hand, $\xcluster$ will be treated as a set of points or as a set of $(\x, \surface)$ pairs depending on context. Note the \free{} points $\xfree$ are shared by every object. We then apply CHSEL to each $\xcluster \cup \xfree$ with a batch of 100 initial transforms centered on the centroid of each $\xcluster$ with uniform orientation. We evaluate the discrepancy (Eq.~\ref{RUMI:eq:cost}) on all registered transforms, assigning the $\xcluster$ with the lowest median discrepancy as the target $k=\targetidx$ to perform \rumiabv{} on, meaning $\xall = \xall_\targetidx \cup \xfree$. This process is depicted in Fig.~\ref{RUMI:fig:select target}.
}

\rev{
When making contact during execution, we assign each contact point the \objcluster{} with the closest centroid. For contact with the target, Algorithm~\ref{RUMI:alg:update} is applied normally. For contact with a non-target ($k \neq \targetidx$), the contact points are added to $\xcluster$ and $\xcluster$ is transformed by the measured change in end effector pose while in contact, essentially assuming sticking contact.
}

\rev{
For planning, we found it advantageous to consider \objcluster{} dynamics and introduce an \textit{obstruction} cost function to penalize moving non-target $\xcluster$ to high information gain regions. The non-target object would be obstructing the robot from observing those points and gaining information. Dynamics is similar to Algorithm~\ref{RUMI:alg:dynamics}, but we replace lines 3-5 with checking for $\x \in \xcluster$ that has the lowest robot \sdf{} value (most inside the robot). Thus, we estimate the displacement of all objects $\displacement_{1..\nclusters}$. The obstruction cost is similar to the information gain cost of Eq.~\ref{RUMI:eq:costinfo}:
\begin{equation}\label{RUMI:eq:costobstruction}
\costobstruction(\displacement_{1..\nclusters,1..\nhorizon}) = \sum_{\x \in \downsample(\bigcup _{i=1}^\nhorizon \bigcup _{k=1, k\neq\targetidx}^\nclusters [\xcluster + \displacement_{k,i} - \displacement_{\targetidx,i}],\resdownsample) } \infofield(\x | \xall)
\end{equation}
where $\xcluster$ is transformed to be the displaced \objcluster{} points in the target object frame. Conveniently, $\costobstruction$ and $\costinfo$ both evaluate information gain in the workspace, with the voxel downsampling $\downsample$ ensuring comparable density of query points, meaning the cost scales can be more easily tuned with respect to each other. 
}

\subsection{Results}

\begin{table}[b]
\centering
\caption{Number of successful trials after 10 runs of active rummaging for pose estimation in different tasks in Fig.~\ref{RUMI:fig:sim rmg init} and Fig.~\ref{RUMI:fig:real rmg init}. Success is defined as the run achieving a minimum $\nll$ below the threshold defined in Tab.~\ref{RUMI:tab:nll}. The most and 1 success below the most successes in each section are bolded.}\label{RUMI:tab:rmg_res_success} 
\begin{tabular}{|l|l|ll|ll|}
\hline
Task        & Ours        & \infoonly   & \reachonly  & \gpvr      & \slide      \\ \hline
sim mug 0   & \textbf{9} & 3           & 7           & 2          & 0           \\
sim mug 1   & \textbf{9}  & 5           & \textbf{8} & 0          & 0           \\
sim mug 2   & \textbf{10} & 6           & \textbf{10} & 7          & 0           \\ \hline
\rowcolor[HTML]{EFEFEF} 
mug total   & \textbf{28} & 14          & 25          & 9          & 0           \\ \hline
sim drill 0 & \textbf{10} & \textbf{9}  & 5           & 1          & 6           \\
sim drill 1 & \textbf{9}  & \textbf{9}  & 7           & 0          & \textbf{10} \\
sim drill 2 & \textbf{7}  & \textbf{7}  & \textbf{7}  & 0          & 0           \\ \hline
\rowcolor[HTML]{EFEFEF} 
drill total & \textbf{26} & \textbf{25} & 19          & 1          & 16          \\ \hline
sim box 0   & \textbf{9}  & \textbf{9}  & \textbf{8}  & 4          & 0           \\
sim box 1   & \textbf{10} & \textbf{9}  & \textbf{10} & 4          & \textbf{10} \\
sim box 2   & \textbf{6}  & \textbf{6}  & 4           & 1          & 1           \\ \hline
\rowcolor[HTML]{EFEFEF} 
box total   & \textbf{25} & \textbf{24} & 22          & 9         & 11          \\ \hline
real mug    & \textbf{7}  & 0           & 4           & 0          & 1           \\ 
real box    & \textbf{7}  & 3           & 0           & 2          & 3           \\ 
\hline
\end{tabular}
\end{table}

\begin{table}[b]
\centering
\caption{Median cumulative $\nll(\xall)$ for 10 runs of active rummaging for pose estimation in different tasks in Fig.~\ref{RUMI:fig:sim rmg init} and Fig.~\ref{RUMI:fig:real rmg init}.
The best and any 5\% within the best are bolded.}\label{RUMI:tab:rmg_res_nll} 

\begin{tabular}{|l|l|ll|ll|}
\hline
Task        & Ours          & \infoonly     & \reachonly & \gpvr          & \slide         \\ \hline
sim mug 0   & \textbf{925}  & 1447          & 1691       & 1691           & 2478           \\
sim mug 1   & \textbf{1447} & 1598          & 1591       & 2252           & 2610           \\
sim mug 2   & 1013          & \textbf{827}  & 1104       & 1239           & 2467           \\ \hline
\rowcolor[HTML]{EFEFEF} 
mug total   & \textbf{3385} & 4444          & 4386       & 5182           & 7555           \\ \hline
sim drill 0 & \textbf{7975} & 10425         & 15401      & 22725          & \textbf{8369}  \\
sim drill 1 & 13750         & 12473         & 21627      & 22835          & \textbf{6049}  \\
sim drill 2 & \textbf{24479} & \textbf{23892} & 27471      & 33419          & 69321          \\ \hline
\rowcolor[HTML]{EFEFEF} 
drill total & \textbf{46204} & \textbf{46790} & 64499      & 78979          & 83739          \\ \hline
sim box 0   & 12744         & 12811         & \textbf{10718} & 24276        & 16527          \\
sim box 1   & 8109          & 7412          & 6596       & 10183           & \textbf{4601}  \\
sim box 2   & \textbf{23923} & 24675         & 43254      & 31196           & \textbf{23157}  \\ \hline
\rowcolor[HTML]{EFEFEF} 
box total   & \textbf{44776} & \textbf{44898} & 60568      & 65655           & \textbf{44285}  \\ \hline
real mug    & \textbf{640}  & 1330           & 946        & 1145           & 4652            \\ 
real box    & 9177  & 8341           & 8762        & 9994            & \textbf{7717}            \\ 
\hline
\end{tabular}
\end{table}

\begin{figure*}[!h]
    \centering
    \includegraphics[width=\linewidth]{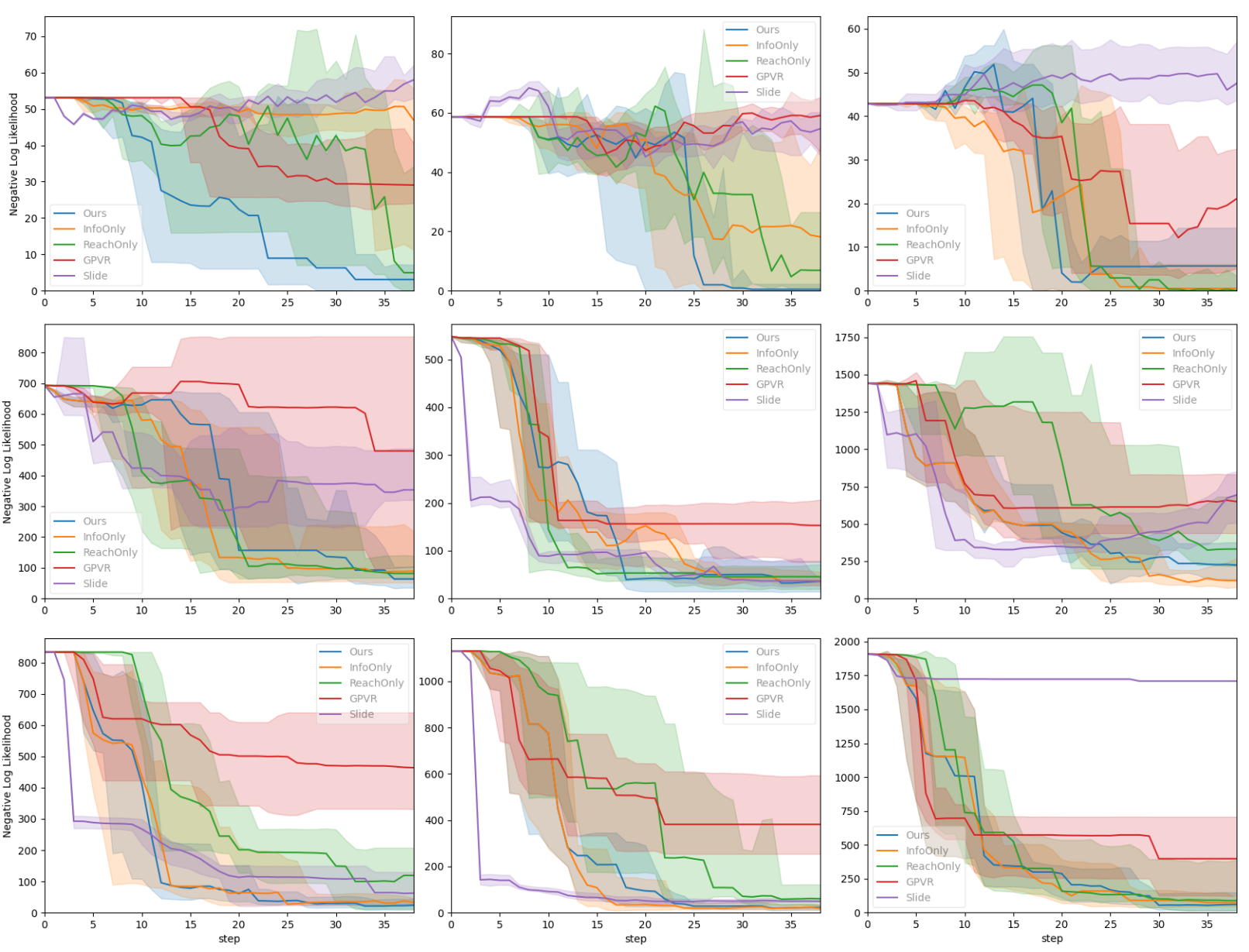}
    \caption{$\nll(\xall)$ after each execution step for simulation tasks depicted in Fig.~\ref{RUMI:fig:sim rmg init}. The median over 10 runs is plotted, with the $25^{th}$ to $75^{th}$ percentile shaded. From top to bottom we have the sim mug, sim drill, and sim box tasks. From left to right we have configuration 0, 1, and 2.}
    \label{RUMI:fig:rmg sim res}
\end{figure*}

\begin{figure}
    \centering
    \includegraphics[width=\linewidth]{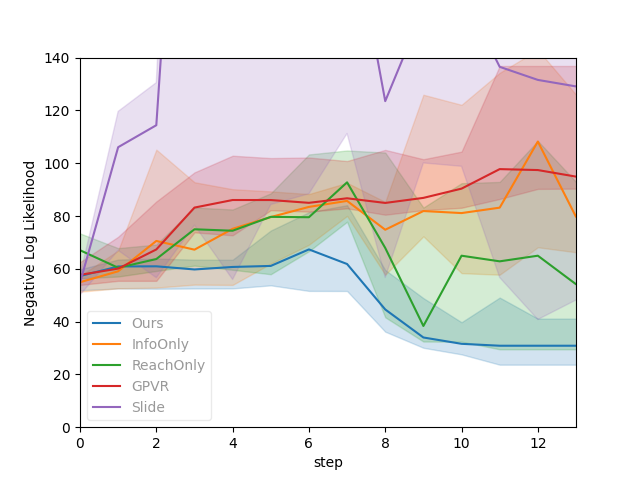}
    \includegraphics[width=\linewidth]{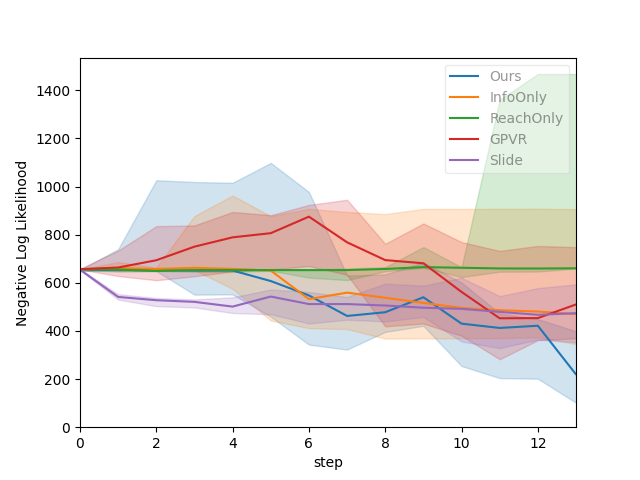}
    \caption{$\nll(\xall)$ after each execution step for the (top) real mug and (bot) real box tasks depicted in Fig.~\ref{RUMI:fig:real rmg init}. The median over 10 runs is plotted, with the $25^{th}$ to $75^{th}$ percentile shaded.}
    \label{RUMI:fig:rmg real res}
    \vspace{-12pt}
\end{figure}

The simulated task results are in Fig.~\ref{RUMI:fig:rmg sim res} and the real task results are in Fig.~\ref{RUMI:fig:rmg real res}.
The number of successful trials out of 10 for each task is compared in Tab.~\ref{RUMI:tab:rmg_res_success}. Additionally, the median over the cumulative $\nll(\xall)$ of each run are in Tab.~\ref{RUMI:tab:rmg_res_nll}. For the sim and real tasks, cumulative $\nll(\xall)$ over time is a good indicator of exploration speed; however, for the box and sim drill tasks, cumulative $\nll(\xall)$ is dominated by the initial search for the first surface points of the object since they do not start with the object in view. Thus, for those tasks it is more a measure of how quickly the different methods make first contact with the object.

We observe that \rumiabv{} is the only method to achieve consistently good performance, if not the most number of successes, across all the sim and real tasks. 
On the sim mug tasks, it also had the lowest cumulative $\nll(\xall)$, meaning it was the most efficient. The ablations show that both $\costinfo$ and $\costreach$ are important for this task, although individually they can also perform well on certain tasks. For example on the sim mug task, \reachonly{} achieved a high number of successes by itself. This was likely due to the handle being close to where the robot needed to push from to increase reachability. However even in this case, adding $\costinfo$ improves efficiency because the mug could be pushed into more reachable regions without contacting the handle. This explains the occasional failures of the \reachonly{} method on the mug tasks. On the drill tasks, pushing the object to be more reachable did not reliably lead to contact that was informative about the pose, and it did much worse than our full method and the \infoonly{} baseline. 

A common failure case for all methods was pushing the object to be outside the robot's reachable region. The performance gain of the full method against the \infoonly{} ablation can be mostly attributed to preventing this. As long as the object was kept within reach and contacts kept being made with the object at different locations, the pose estimation was gradually improved. This is illustrated in \reachonly{}'s performance on the sim box tasks, where it is one of the slowest methods to reduce $\nll$, but was still able to achieve a relatively high number of successful trials.

The \slide{} baseline exhibited behavior that in some ways was the opposite of \reachonly{}'s. It always pushed the object away from the robot, and it became a race of it gathering enough pose-identifying information from those contacts before the object moved out of reach. On the real robot, sometimes it did not register that a contact was made and would continue pushing forward. This strategy's success was highly configuration-dependent, seen in Tab.~\ref{RUMI:tab:rmg_res_success}, where it can either achieve reliable success (since there is only randomness in the sliding direction), or no success. 
This strategy however does often lead to it being the quickest method to make contact with the object, giving it low cumulative $\nll(\xall)$ for the sim drill and box tasks.

\begin{figure}
    \centering
    \includegraphics[width=0.49\linewidth]{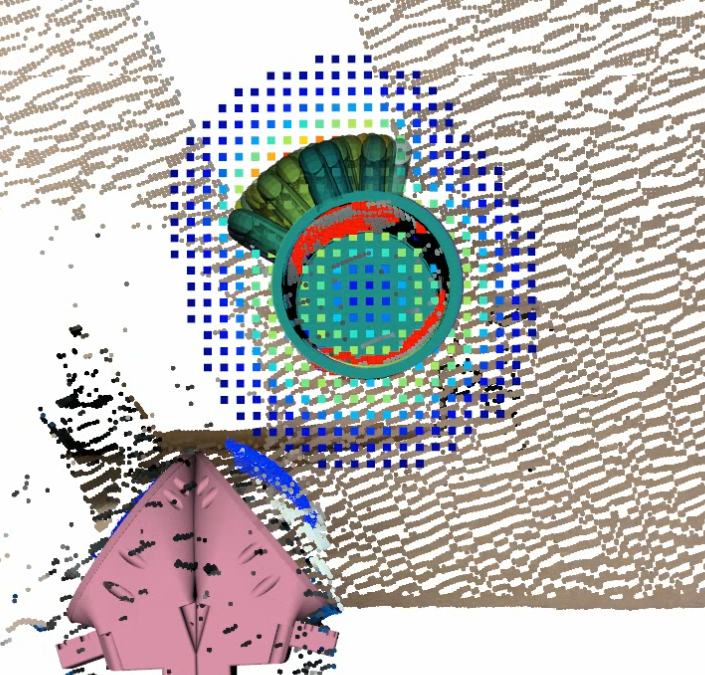}
    \includegraphics[width=0.49\linewidth]{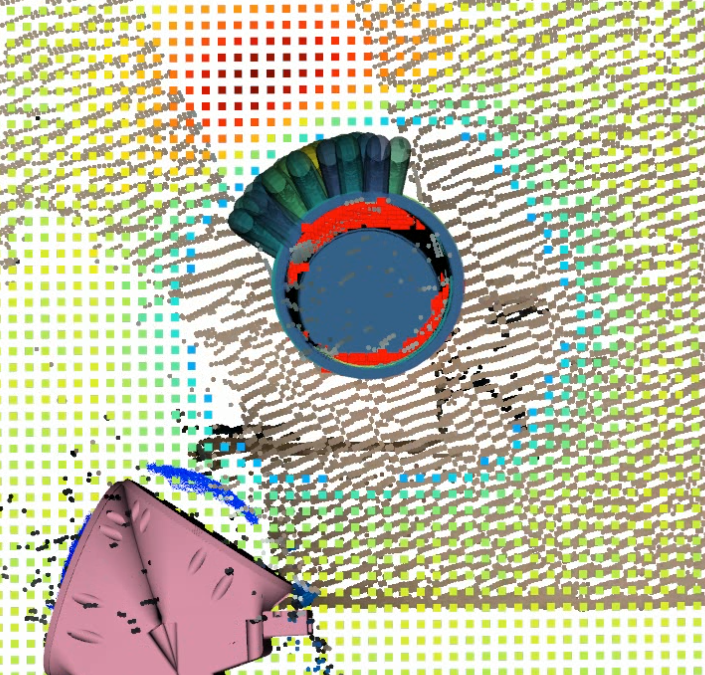}
    \caption{Comparison of the fields to plan over evaluated at each $\x \in \workspace$ for (left) our method using $\infofield(\x|\xall)$ against the (right) \gpvr{} baseline using $\var(\x|\xall)$. $\x$ with too low $\infofield$ or $\var$ are omitted.}
    \label{RUMI:fig:gpvr vs ours}
\end{figure}

The \gpvr{} baseline's performance can be compared against the \infoonly{} ablation's, as neither have an explicit cost for avoiding the object from being pushed out. As seen in Fig.~\ref{RUMI:fig:gpvr vs ours}, the highest $\var(\x|\xall)$, even when given points augmented using shape information, does not match where intuitively information about the shape might be held. A similar problem was present in the drill tasks, where \gpvr{} does very poorly because the task requires making multiple contacts close together, such as on either side of the drill head. Upon making contact with one side, the proximity of observed surface points lowers the GP variance around it, placing high cost on visiting the other side or the front of the drill, which was necessary to estimate its pose. This suggests that augmenting points is not a satisfactory way of conditioning on known object shape. 

\begin{figure}
    \centering
    \includegraphics[width=\linewidth]{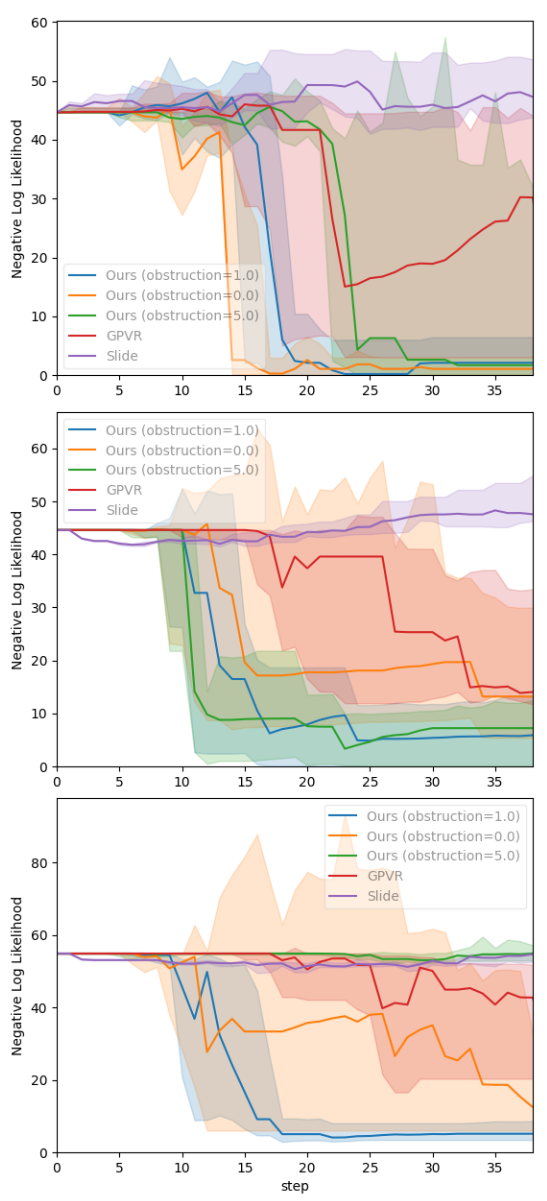}
    \caption{\rev{$\nll(\xall)$ after each execution step for clutter tasks depicted in Fig.~\ref{RUMI:fig:clutter start}. The median over 10 runs is plotted, with the $25^{th}$ to $75^{th}$ percentile shaded. From top to bottom we have Clutter Easy, Clutter Medium, and Clutter Hard.}}
    \label{RUMI:fig:clutter res}
\end{figure}

\rev{
The clutter task results are in Fig.~\ref{RUMI:fig:clutter res} where we see that our method performs the best on all tasks. Different scales for $\costobstruction$ were considered, with scale=1 relative to information gain performing the best. Having scale=0 led to pushing the other objects into the target object, while scale=5 led to avoiding pushing the non-target objects at all, making it impossible to do well on Clutter Hard.
}

\subsection{Runtime Comparison}
We also recorded the average computation time per step in the sim mug task and sim box task to highlight \rumiabv{}'s computational efficiency in Tab.~\ref{RUMI:tab:rmg runtime}. 
Caching $p(\Sx|\xall)$ and $\infofield(\x|\xall)$, and the dynamics were processes shared by all methods.
All methods were implemented in PyTorch
and accelerated by running on a modern computer with a NVIDIA RTX 4090 GPU.
Computing $p(\Sx|\xall)$ and $\infofield(\x|\xall)$ for $\x \in \workspace$ took 0.061s per step, while evaluating our cost inside the MPC took 0.178s for the sim mug. 
The time was dominated by evaluating $\configdynest$ because it is stochastic and so benefited from sampling multiple state rollouts, in addition to dividing each step into 4 sequentially applied mini steps to avoid over-penetration. The GP fitting process also included caching $\var(\x|\xall) \ \forall \x \in \workspace$ in a voxel grid to speed up inference inside the cost. 

We considered how well the methods scale to the full 3D sim box task. The main challenge was the increased $\workspace$ size, with approximately $2.24$ times more total points. Caching $p(\Sx|\xall)$ and $\infofield(\x|\xall)$ slowed down to 0.387s, or an increase of $6.3$ times, while our cost evaluation increased around 3 times to 0.556s. The \gpvr{} cost run time scaled well because we were down sampling the workspace by 7 times the resolution for fitting the GP's free space. 

Reducing the step size to no longer require dynamics mini steps, or using an alternative dynamics function would effectively improve the whole method's efficiency. Currently, \rumiabv{} can be run at around 1Hz, which was more than sufficient for quasi-static rummaging.

\begin{table}[b]
\centering
\caption{Run time for different processes of our method and the \gpvr{} baseline per execution step of the planar sim mug 0 and the 3D sim box 0 task across 10 runs. Standard deviation is in parenthesis.}
\label{RUMI:tab:rmg runtime}
\begin{tabular}{|l|ll|}
\hline
\multirow{2}{*}{process}                     & \multicolumn{2}{l|}{average time per step (s)} \\ \cline{2-3} 
                                             & sim mug                & sim box             \\ \hline
cache $p(\Sx|\xall)$, $\infofield(\x|\xall)$ & 0.061 (0.001)          & 0.387 (0.002)         \\
dynamics                                     & 0.724 (0.012)          & 1.242 (0.054)         \\
our cost lookup                              & 0.178 (0.002)          & 0.556 (0.021)         \\
GP fit                                       & 1.870 (0.066)          & 2.541 (0.061)         \\
GP cost lookup                               & 0.116 (0.002)          & 0.184 (0.004)         \\ \hline
\end{tabular}
\end{table}

\section{Discussion and Future Work}

\subsection{Object Clutter}
\rev{
Qualitatively, the robot pushed multiple obstructing non-target objects to clear a path to the back of the mug for Clutter Hard, necessary for this task. This is a true interactive perception problem, rather than one arising from the lack of reliable and mobile vision.
The good performance of \rumiabv{} on clutter tasks suggests potential for future practical applications. Only the target object mesh was known, with the number, placement, and shapes of the clutter objects being unknown. Critical to this success was the accurate initial clustering of the camera observation into \objcluster{}s, at least in separating the target from other objects, which may be more difficult in real environments with more noise. This required the objects to start sufficiently separated from each other, as well as for the objects to geometrically be sufficiently different from the target object. Non-geometric features such as color could be incorporated in this initial clustering step in future work.
}

\rev{
Future work could also explore the total information gain provided by all \objcluster{}s; however, an obstacle to that is the fact that the information gain is unnormalized, so $\infofield(\x|\xcluster \cup \xfree)$ cannot be trivially summed for $k=1..\nclusters$.
}

\subsection{Limitations}
\rev{
In addition to requiring the target object to be geometrically distinct from the rest of the environment,
another limitation is the need to avoid object toppling while exploring the environment. This places restrictions on object geometry and requires slow robot movement. Since after the initial visual observation we assume no reliable vision, we would have no way to sense and respond to object toppling. Incorporating mobile vision could be explored to respond to more diverse environments.
}

\subsection{Effects of Approximations}
\rev{
We used several approximations in the formulation of our method. Although it is difficult to quantitatively evaluate the effect of each, we provide a qualitative discussion to provide more insight into their effects. First, the use of reverse KL instead of forward KL in Eq.~\ref{RUMI:eq:revkl}. We note that the difference is only significant in regions where $\infoklrev$ and $\infoklfor$ are high, and then only in relative magnitude. They predominantly agree categorically on which regions have near 0 information gain or significant information gain, so we expect this approximation to not affect exploration performance. Secondly, there is the dropping of normalizing constants in Eq.~\ref{RUMI:eq:revkl normalizing constant}. Their contribution is controlled by $\peakiness$; with a lower $\peakiness$ the approximation accuracy goes down because the ignored term becomes relatively larger, but having too high $\peakiness$ leads to particle degeneracy in the particle filter approximation of the pose posterior. Finally, there is the particle filter approximation itself. From Fig.~\ref{RUMI:fig:numparticles} we qualitatively see the effect of increasing $\numparticles$, which is to reduce the variance of the approximation. Beyond $\numparticles=100$, we found little practical downstream benefit in exploration performance.
}

\subsection{Unknown Object Shape}
The last point of improvement is to relax our knowledge of the object from having its SDF to just having a class label. One naive approach is to use a template SDF for each object class and absorb the SDF uncertainty into the sensor model $\p(\Sx|\sdf(\T\x))$. However, this fails with object classes that have high geometric variation. One possible approach would be to extend the pose posterior particle filter to also represent object shape, such that each particle is both a pose and a shape. The shape could be parameterized by recent advances in 3D representations such as the Deformed Implicit Field~\cite{lee2022uncertainty} that allows shape editing by constraining on surface points.

\section{Conclusion}
We presented \rumiabv{}, an active exploration method based on the mutual information between a movable target object's uncertain pose and the robot trajectory. It maintains an explicit belief over the object pose using a particle filter, updating it with observed point clouds augmented with semantics, such as whether a point is in free space or on the object surface. Given object SDF, we formulated an information gain cost function evaluating the expected KL divergence between the pose distribution before and after executing a robot trajectory. In addition, we implemented a reachability cost function and showed that it was important to prevent pushing the object outside the robot's reachable region. Through comparison with baselines in real and simulated experiments, we showed that \rumiabv{} could effectively and efficiently condition on object shape to explore and estimate object pose. \rev{Additionally, we demonstrated potential for practical applications with tasks where an unknown number of unknown movable objects were cluttered around the target object.}


\footnotesize
\bibliographystyle{plainnat}
\bibliography{references}

\end{document}